\documentclass{article}

\usepackage[final,sglblindworkshop]{neurips_2026}
\workshoptitle{Interpretability for Discovery}

\usepackage[utf8]{inputenc}
\usepackage[T1]{fontenc}
\usepackage{microtype}
\usepackage{setspace}
\usepackage{siunitx}
\usepackage{amsmath, amssymb, mathtools}
\usepackage{graphicx}
\usepackage{subcaption}
\usepackage{multirow}
\usepackage{float}
\usepackage{algorithm}
\usepackage{algpseudocode}
\usepackage{amsthm}
\usepackage{makecell}
\usepackage{rotating}
\usepackage{array}
\usepackage{colortbl}
\usepackage[hyphens]{url}
\usepackage{xcolor}
\usepackage{wrapfig}
\usepackage{csquotes}
\usepackage{booktabs}
\usepackage{xspace}
\usepackage{hyperref}
\hypersetup{
  colorlinks=true,
  linkcolor=blue!50!black,
  citecolor=blue!50!black,
  urlcolor=blue!60!black
}
\usepackage[capitalise,noabbrev]{cleveref}

\newcommand{\KL}{$\mathrm{KL}$ }
\newcommand{\acronymcite}[2]{\\[-0.3ex]\tiny[#2]}

\newcolumntype{R}{>{\begin{turn}{90}}l<{\end{turn}}}

\title{$\textsc{XFactors}$: Disentangled Information Bottleneck\\via Contrastive Supervision}

\author{%
  {\normalfont\normalsize Alexandre Myara$^*$ \quad Nicolas Bourriez$^*$ \quad Thomas Boyer$^*$}\\[0.15em]
  {\normalfont\normalsize Ihab Bendidi \quad Thomas Lemercier \quad Auguste Genovesio$^\dagger$}\\[0.15em]
  {\normalfont\small \'Ecole Normale Sup\'erieure (IBENS), Paris, France}\\[0.15em]
  {\normalfont\footnotesize \texttt{alexandre.myara@ens.fr} \quad \texttt{nicolas.bourriez@ens.fr} \quad \texttt{thomas.boyer@ens.fr}}\\
  {\normalfont\footnotesize \texttt{ihab.bendidi@ens.fr} \quad \texttt{thomas.lemercier@ens.fr} \quad \texttt{auguste.genovesio@ens.fr}}
}

\begin{document}

\maketitle
\vspace{-1.5em}
\enlargethispage{2\baselineskip}

\begingroup
  \renewcommand{\thefootnote}{\fnsymbol{footnote}}
  \footnotetext[1]{\raggedright These authors contributed equally.}
  \footnotetext[2]{\raggedright Corresponding author: Auguste Genovesio.}
\endgroup

\begin{abstract}
  Modern deep networks often learn features predictive of semantic attributes, but these factors are typically distributed across the representation rather than exposed as stable, addressable, testable variables. We study a practical version of disentanglement in which only a subset of factors is known, annotated, or actionable, the rest of the variation being unannotated, irrelevant to the downstream question, or too numerous to isolate exhaustively.
  We introduce \textsc{XFactors}, a weakly-supervised VAE that makes selected factors directly interpretable through an explicit latent interface: the representation is decomposed into factor-specific subspaces $\mathcal{T}_1,\ldots,\mathcal{T}_K$ and a residual subspace $\mathcal{S}$, contrastive objectives align each selected factor with its assigned $\mathcal{T}_i$, and reconstruction and KL regularization retain non-targeted variation in $\mathcal{S}$ subject to the VAE bottleneck. The non-adversarial objective scales to multiple target factors by assigning one contrastive signal per block. With constant hyperparameters across benchmarks, \textsc{XFactors} obtains strong disentanglement scores, supports factor swapping by latent replacement, scales with residual capacity, and provides qualitative proof-of-concept results on real-world CelebA and JUMP Cell Painting data, where known biological and technical variables become directly interpretable.\footnote{Our code is available at \url{https://osf.io/65kbh/overview?view_only=57e6352f8bcd4a42a39f3b68ab0b1933}.}
\end{abstract}

\begin{figure}[!ht]
    \centering
    \begin{subfigure}[b]{0.368\linewidth}
        \centering
        \includegraphics[width=\linewidth]{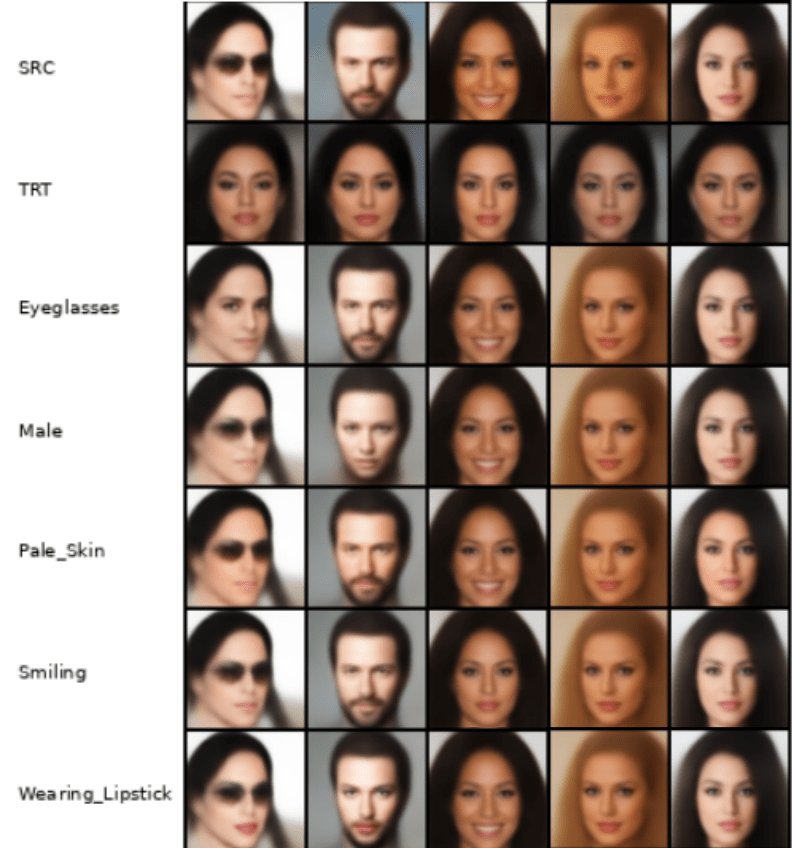}
        \caption{CelebA (5 target attributes)}
        \label{fig:gens_celeba}
    \end{subfigure}\hfill
    \begin{subfigure}[b]{0.592\linewidth}
        \centering
        \includegraphics[width=\linewidth]{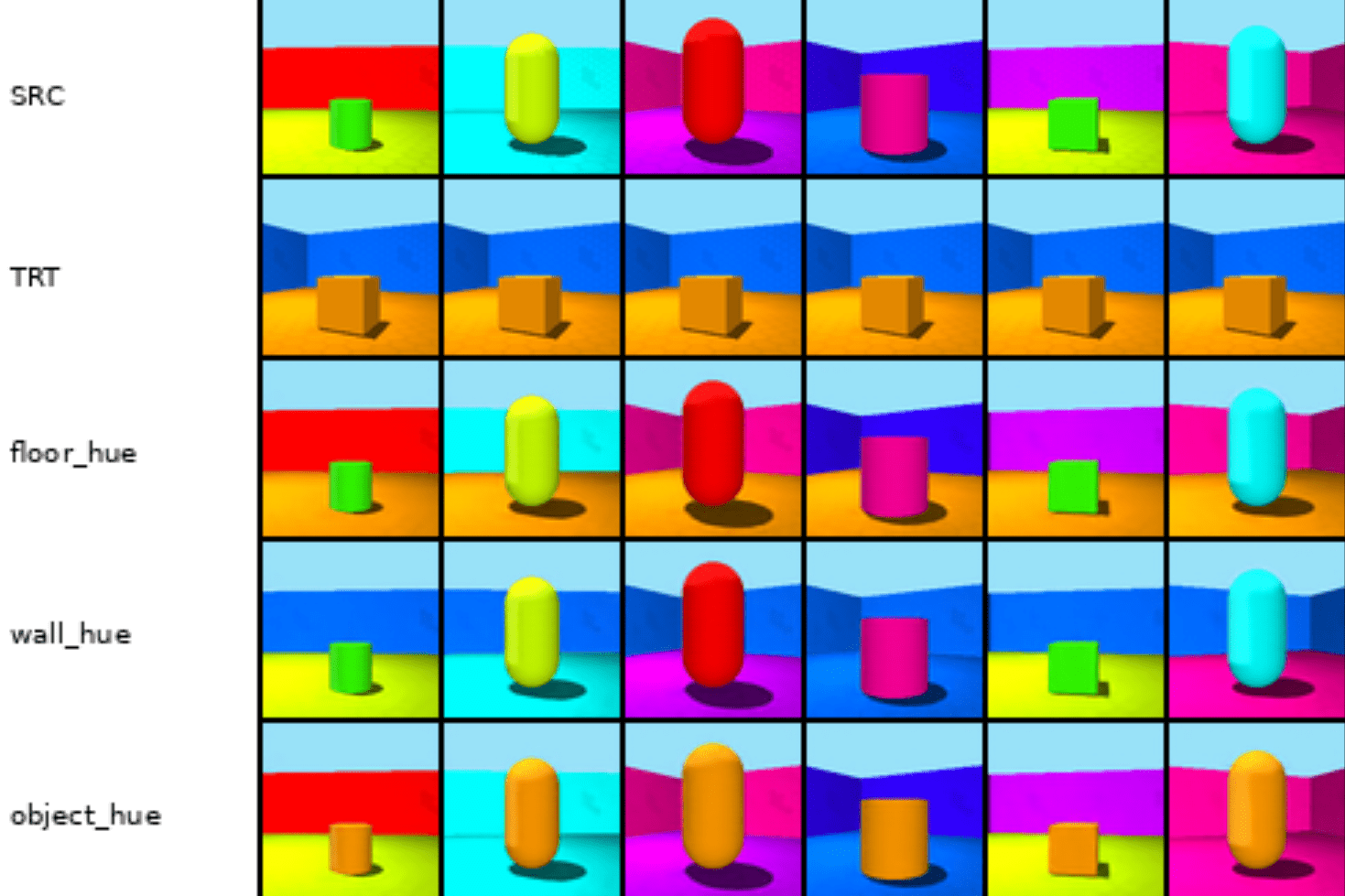}
        \caption{Shapes3D (5 target factors)}
        \label{fig:gens_shapes}
    \end{subfigure}
    \caption{\small \textbf{The \textsc{XFactors} latent interface: factor swapping by latent replacement.} For each row (source, top) we replace a single target block $T_i$ of the source code by the corresponding block of the target image (TRT) and decode; every other latent component, including the residual code $z_s$, is kept fixed, so each column edits one named factor. Source and target reconstructions are shown for reference.}
    \label{fig:teaser}
\end{figure}

\section{Introduction}\label{sec:introduction}

Disentangled representation learning seeks latent variables that expose semantic factors of variation in separate, interpretable components \citep{bengio2014representationlearningreviewnew}, supporting counterfactual simulation, out-of-domain analysis, and generalization to unseen factor combinations \citep{steenbrugge2018improvinggeneralizationabstractreasoning, sauer2021counterfactualgenerativenetworks, pan2020disentangledinformationbottleneck}. Yet strong predictive or generative performance does not guarantee that the learned factors are accessible: semantic variations are usually distributed across many dimensions rather than exposed as stable variables that a domain expert can inspect, replace, or confirm.

Fully disentangling all latent causes is rarely the realistic goal. In real-world datasets, only a few factors are known, actionable, annotated, or scientifically relevant, while the rest may be unknown, expensive to annotate, irrelevant to the downstream question, or too numerous to isolate; models also reproduce training-set correlations, including spurious ones \citep{Geirhos_2020}. Unsupervised disentanglement therefore requires strong inductive biases or favorable data assumptions, and leaves the user searching post-hoc over latent directions before any coordinate can be tied to a semantic factor \citep{locatello2018challengingcommonassumptionsunsupervised, träuble2021disentangledrepresentationslearnedcorrelated}. Classifier-based supervision is more direct, but decision boundaries are sensitive to adversarial perturbations \citep{szegedy2014intriguingpropertiesneuralnetworks, augustin2022diffusionvisualcounterfactualexplanations} and do not by themselves provide a generative handle for intervening on one factor while keeping the rest fixed.

This motivates a targeted form of disentanglement, where the factors to make addressable are specified in advance and the remaining variation is retained without being forced into the selected factor spaces. Such a setting matches many controlled datasets with partial labels and incomplete combinatorial coverage, such as screening datasets covering drug compounds across a limited set of cell lines, where exhaustive combinations are prohibitively expensive \citep{Lotfollahi2023}.

In this work, we present \textsc{XFactors}, a weakly-supervised method that gives explicit control over which factors are disentangled among the possibly unknown latent variations of the data. It requires labels for those factors only, and avoids encoding spurious relations between them or with any other non-targeted or unknown factor. The resulting representation is not merely predictive of the selected factors, but \emph{addressable}: a block $T_i$ can be compared, replaced, or intervened on while the rest of the latent code stays fixed (\cref{fig:teaser}), enabling counterfactual generation, factor swapping, nuisance-factor analysis, testable hypotheses about named factors, and dataset debugging. The residual space $\mathcal{S}$ absorbs non-targeted variation such as morphology, identity, pose, background, or technical effects, reducing the pressure to encode it in the supervised blocks.

Our contributions are as follows:
\begin{itemize}\itemsep1pt \parskip0pt
    \item We formulate \emph{targeted partial disentanglement}: selecting a subset of labeled factors to control while fully retaining non-targeted variation in a residual subspace.
    \item We propose \textsc{XFactors}, a VAE with factor-specific latent blocks supervised by contrastive objectives and a KL-regularized residual block, avoiding classifier-based and adversarial disentanglement.
    \item We show that it achieves strong disentanglement on standard benchmarks, supports factor swapping, scales with residual capacity, and exposes interpretable factors on real-world CelebA and JUMP Cell Painting data.
\end{itemize}

\section{Methods}\label{sec:methods}\label{sec:archi}

\textsc{XFactors} assigns each labeled factor to its own block $\mathcal{T}_i$ and leaves non-targeted information to $\mathcal{S}$, subject to the VAE bottleneck. The objective is a direct minimization problem: no adversarial training, no auxiliary classifier as the main disentanglement mechanism, and one contrastive loss per $\mathcal{T}_i$, so it scales to multiple target factors.

\paragraph{X-DisenIB.}
Given a dataset $(X,Y)$, we denote $(y_{f_1},\ldots,y_{f_K})$ the annotated factors of samples in $X$ with $K$ the number of factors we aim to disentangle, $\mathcal{Z}$ the latent space, and $Z\subset\mathcal{Z}$ our latent set. We consider subspaces $\mathcal{S}$ and $\mathcal{T}_1,\ldots,\mathcal{T}_K$ of $\mathcal{Z}$ such that the latent space decomposes as a direct sum:
\begin{equation}
    \mathcal{Z} = \mathcal{S} \bigoplus_{i=1}^K \mathcal{T}_i
    \label{eq:directsum}
\end{equation}
The decomposition in \cref{eq:directsum} lets us encourage the $i^{\text{th}}$ factor to be represented in $\mathcal{T}_i$ by promoting high information between $T_i \subset \mathcal{T}_i$ and its label $y_{f_i}$. We accordingly adapt the Disentangled Information Bottleneck of \citet{pan2020disentangledinformationbottleneck} to this decomposition:
\begin{equation}
    \mathcal{L}_{\textrm{X-IB}} = -\sum_{i=1}^K \mathcal{I}(T_i;y_{f_i}) - \mathcal{I}(X;(S,Y)) + \mathcal{I}(T;S)
    \label{eq:xib}
\end{equation}
The last term is an idealized motivation for keeping target and residual subspaces distinct: in the implemented objective, that separation comes from the latent partition, the contrastive block losses, the reconstruction bottleneck, and KL regularization rather than from an explicit estimator of $\mathcal{I}(T;S)$. To optimize the target-factor part of \cref{eq:xib}, we use the supervised InfoNCE loss of Supp.~Eq.~\eqref{eq:supervised_infonce} to promote clustering in $\mathcal{T}_i$ according to $y_{f_i}$. This block geometry is useful for recognition, but also for editing and swapping, because the supervised factor is localized in a known part of the representation, while non-targeted variation is encoded in $\mathcal{S}$ without any explicit clustering objective.

A Gaussian \KL\xspace regularizer helps organize latent spaces even without labels \citep{burgess2018understandingdisentanglingbetavae}, so we impose Gaussian priors on $\mathcal{S}$ with $\mathcal{L}_{\textrm{KL}}^{\mathcal{S}}$ and on the aggregated factor subspace $\bigoplus_{i=1}^K \mathcal{T}_i$ with $\mathcal{L}_{\textrm{KL}}^{\mathcal{T}}$. To retain information about $\mathbf{x}$ in $\mathbf{z}\in\mathcal{Z}$ subject to this bottleneck, we use a reconstruction $L^2$ loss $\mathcal{L_\text{reco}}$ as a proxy.

\paragraph{\textsc{XFactors} losses.}
Motivated by \cref{eq:xib}, our implemented objective is:
\begin{equation}
    \mathcal{L} = \mathcal{L}_{\text{reco}}
    + \beta_s \cdot \mathcal{L}_{\textrm{KL}}^{\mathcal{S}}
    + \beta_t \cdot \mathcal{L}_{\textrm{KL}}^{\mathcal{T}}
    + \sum_{i=1}^K\lambda_i \cdot \mathcal{L}_{\text{InfoNCE}}^{(i)}
    \label{eq:xfactors_loss}
\end{equation}

\begin{figure}[t]
    \centering
    \includegraphics[width=0.82\linewidth]{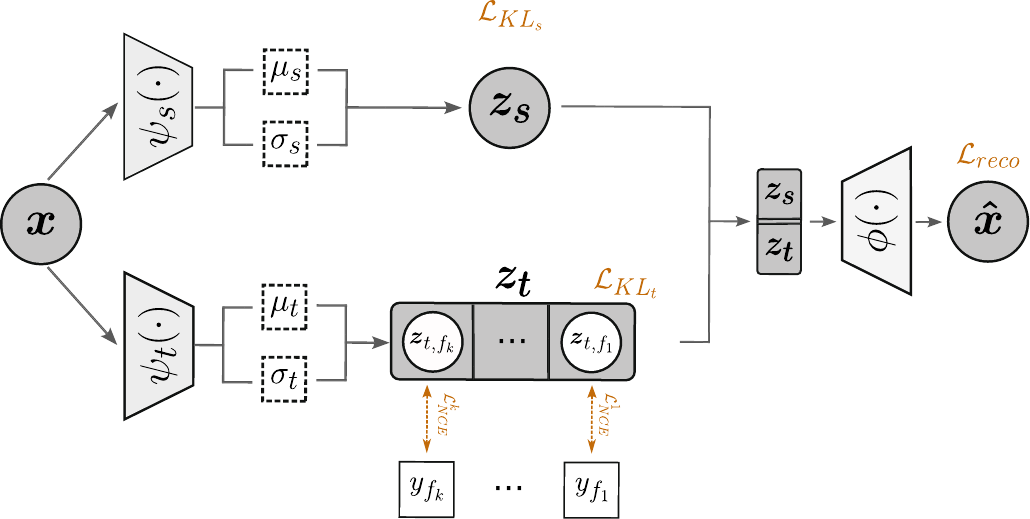}
    \caption{\small \textbf{Our architecture}. \textsc{XFactors} processes the input $\boldsymbol{x}$ using two parallel encoders: $\psi_s(\cdot)$, which captures the residual information in the latent code $\boldsymbol{z}_s$, and $\psi_t(\cdot)$, which encodes the factors of interest in $\boldsymbol{z}_t$. The factor latent $\boldsymbol{z}_t$ is explicitly disentangled by aligning specific subspaces $\boldsymbol{z}_{t,f_i}$ with their corresponding ground-truth labels $y_{f_i}$ via InfoNCE objectives ($\mathcal{L}_{\text{NCE}}$). The latent spaces are regularized using KL-divergence terms ($\mathcal{L}_{KL_s}$, $\mathcal{L}_{KL_t}$), and the decoder $\phi(\cdot)$ reconstructs the input $\boldsymbol{\hat{x}}$ from the concatenated latent representation $[\boldsymbol{z}_s \| \boldsymbol{z}_t]$.}
    \label{fig:archi}
\end{figure}

To train \cref{eq:xfactors_loss}, we use two variational encoders $\psi_s$ and $\psi_t$ to produce $\mathcal{S}$ and $\bigoplus_{i=1}^K\mathcal{T}_i$ (resp.). We concatenate $z_s\in\mathcal{S}$ and $z_t \in \bigoplus_{i=1}^K\mathcal{T}_i$ as in \cref{fig:archi}, and decode the resulting $z\in\mathcal{Z}$ with $\phi$.

\section{Experiments}
\label{sec:experiments}

\paragraph{Datasets}\label{par:datasets}
We first consider four fully disentangled datasets: 3DShapes \citep{3dshapes18}, Cars3D \citep{cars3D}, dSprites \citep{dsprites17}, and MPI3D \citep{NEURIPS2019_d97d404b}, each providing 3--7 annotated factors with full combinatorial coverage. We associate each supervised factor to a subspace $\mathcal{T}_i$, except one factor left for $S$ without explicit supervision. We also consider CelebA \citep{liu2015faceattributes}, whose 40 binary facial attributes are correlated and do not form a fully disentangled generative process: we target 5 attributes in $\mathcal{T}_1,\ldots,\mathcal{T}_5$ and leave the remaining variation unconstrained in $\mathcal{S}$.

\paragraph{Hyperparameters}
Except in Supp.~\ref{sec:scaling_laws}, we use a single network architecture and set of hyperparameters across all datasets: $\lambda_{\mathrm{NCE}}=0.5$, $\text{dim}_{\mathcal{T}_i}=2$, $\text{dim}_\mathcal{S}=126$, $\beta_s=100$, $\beta_t=100$. This single configuration yields stable training and consistent performance without dataset-specific tuning.
\subsection{Disentanglement metrics}\label{sec:metrics}
We report two common disentanglement scores in \cref{tab:disentanglement-metrics}: FactorVAE \citep{kim2019disentanglingfactorising} and DCI\footnote{Since compactness $C$ and informativeness $I$ are rarely reported by prior baselines, we defer the full DCI triplets to Supp.~Table~\ref{tab:supp_xfactors_dci_triplets}} \citep{DCI}. On the fully disentangled datasets, \textsc{XFactors} obtains the highest reported scores. CelebA scores (Supp.~Table~\ref{tab:celeba_metrics}) need more caution, as methods differ in supervision and in the number of attributes targeted: we supervise only five attributes and leave the remaining facial variation in $\mathcal{S}$, so those scores assess whether the selected attributes are localized, not whether all facial causes were recovered.

\begin{table}[H]
  \centering
  \fontsize{8}{9.5}\selectfont
  \setlength{\tabcolsep}{3pt}
  \caption{\footnotesize Disentanglement metrics (\underline{F}actor\underline{VAE} and DCI) on classical disentangled datasets; best value in \textbf{bold}, details in Supp.~\ref{sec:supp_disentanglement_metrics_details}. $|\mathcal{S}|=i$ is the number of annotated factors grouped in $\mathcal{S}$ instead of being isolated in $\mathcal{T}$, the whole of $\mathcal{S}$ then counting as a single block factor $\mathbf{s}$: $|\mathcal{S}|=1$ matches prior settings, while $|\mathcal{S}|=3$ tests whether disentanglement survives grouping several factors, as in real-world applications.}
  \label{tab:disentanglement-metrics}
  \hspace*{-8mm}
  \begin{tabular}{cc!{\color{gray!50}\vrule width 0.3pt}ccc|cccccc}
    \toprule
                                                                                                   &                                                        & \multicolumn{3}{|c|}{Unsupervised methods} & \multicolumn{5}{c}{Weakly supervised methods} \\
    \midrule
    \textbf{Dataset}                                                                               & \makecell{\textbf{Metric}                                                                                                                           \\($\uparrow$)} &
    \makecell{FactorVAE\acronymcite{kim2019disentanglingfactorising}{K\&M+18}}                     &
    \makecell{$\beta$-TCVAE\acronymcite{chen2018isolatingsourcesdisentanglementvariational}{C+18}} &
    \makecell{EncDiff\acronymcite{yang2024diffusionmodelcrossattention}{Y+24}}                     &
    \makecell{DSD\acronymcite{feng2018dualswapdisentangling}{F+18}}                                &
    \makecell{Ada-ML-VAE\acronymcite{locatello20a}{L+20}}                                          &
    \makecell{Ada-GVAE\acronymcite{locatello20a}{L+20}}                                            &
    \makecell{SW-VAE(e)\acronymcite{zhu2022swvae}{Z+22}}                                           &
    \makecell{\textsc{XFactors}\\{\tiny ($|S|=1$, ours)}}                    &
    \makecell{\textsc{XFactors}\\{\tiny ($|S|=3$, ours)}}                                                                                                                                                                          \\
    \midrule
    \multirow{2}{*}{Shapes3D}
                                                                                                   & FVAE
                                                                                                   & $.840{\scriptstyle \pm .066}$
                                                                                                   & $.873{\scriptstyle \pm .074}$
                                                                                                   & $.999{\scriptstyle \pm .000}$
                                                                                                   & $.997$
                                                                                                   & $.996$
                                                                                                   & $\mathbf{1.{\scriptstyle 000}}$
                                                                                                   & $.998$
                                                                                                   & $\mathbf{1.{\scriptstyle 000}}{\scriptstyle \pm .000}$
                                                                                                   & $\mathbf{1.{\scriptstyle 000}}$
    \\
                                                                                                   & DCI
                                                                                                   & $.611{\scriptstyle \pm .082}$
                                                                                                   & $.613{\scriptstyle \pm .114}$
                                                                                                   & $.969{\scriptstyle \pm .030}$
                                                                                                   & $.902$
                                                                                                   & $.940$
                                                                                                   & $.946$
                                                                                                   & $.920$
                                                                                                   & $\mathbf{1.{\scriptstyle 000}}{\scriptstyle \pm .000}$
                                                                                                   & $\mathbf{1.{\scriptstyle 000}}$
    \\
    \midrule
    \multirow{2}{*}{MPI3D}
                                                                                                   & FVAE
                                                                                                   & $.152{\scriptstyle \pm .025}$
                                                                                                   & $.179{\scriptstyle \pm .017}$
                                                                                                   & $.872{\scriptstyle \pm .049}$
                                                                                                   & $.510$
                                                                                                   & $.476$
                                                                                                   & $.621$
                                                                                                   & $.520$
                                                                                                   & $.\mathbf{978}{\scriptstyle \pm .001}$
                                                                                                   & $\mathbf{.999}$
    \\
                                                                                                   & DCI
                                                                                                   & $.240{\scriptstyle \pm .051}$
                                                                                                   & $.237{\scriptstyle \pm .056}$
                                                                                                   & $.685{\scriptstyle \pm .044}$
                                                                                                   & $.445$
                                                                                                   & $.285$
                                                                                                   & $.401$
                                                                                                   & $.554$
                                                                                                   & $.\mathbf{949}{\scriptstyle \pm .000}$
                                                                                                   & $\mathbf{.980}$
    \\
    \midrule
    \multirow{2}{*}{dSprites}
                                                                                                   & FVAE
                                                                                                   & $\approx.82$
                                                                                                   & --
                                                                                                   & --
                                                                                                   & $.912$
                                                                                                   & $.701$
                                                                                                   & $.847$
                                                                                                   & $.891$
                                                                                                   & $\mathbf{.952}{\scriptstyle \pm .003}$
                                                                                                   & $\mathbf{.990}$
    \\
                                                                                                   & DCI
                                                                                                   & --
                                                                                                   & --
                                                                                                   & --
                                                                                                   & $.602$
                                                                                                   & $.294$
                                                                                                   & $.479$
                                                                                                   & $.685$
                                                                                                   & $.\mathbf{909}{\scriptstyle \pm .001}$
                                                                                                   & $\mathbf{.824}$
    \\
    \midrule
    \multirow{2}{*}{cars3D}
                                                                                                   & FVAE
                                                                                                   & $.906{\scriptstyle \pm .052}$
                                                                                                   & $.855{\scriptstyle \pm .082}$
                                                                                                   & $.773{\scriptstyle \pm .060}$
                                                                                                   & --
                                                                                                   & $.874$
                                                                                                   & $.902$
                                                                                                   & --
                                                                                                   & $.\mathbf{948}{\scriptstyle \pm .002}$
                                                                                                   & $\mathbf{1.{\scriptstyle 000}}$
    \\
                                                                                                   & DCI
                                                                                                   & $.161{\scriptstyle \pm .019}$
                                                                                                   & $.140{\scriptstyle \pm .019}$
                                                                                                   & $.279{\scriptstyle \pm .022}$
                                                                                                   & --
                                                                                                   & $.456$
                                                                                                   & $.540$
                                                                                                   & --
                                                                                                   & $.\mathbf{802}{\scriptstyle \pm .004}$
                                                                                                   & $.\mathbf{885}$
    \\
    \bottomrule
  \end{tabular}
\end{table}

\subsection{Latent structure}
\label{sec:latent_plots}
\begin{figure}[t]
  \centering
  \includegraphics[width=0.32\linewidth]{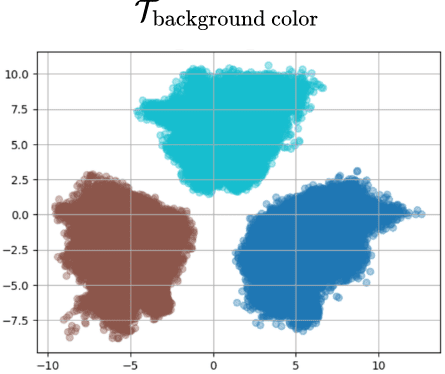}\hfill
  \includegraphics[width=0.32\linewidth]{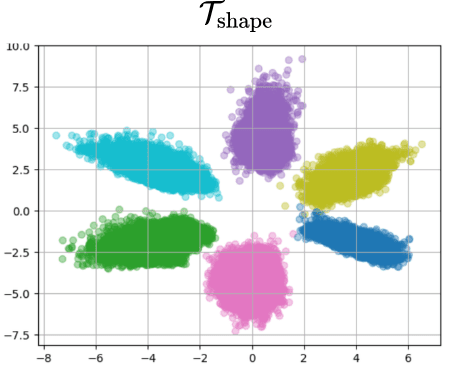}\hfill
  \includegraphics[width=0.32\linewidth]{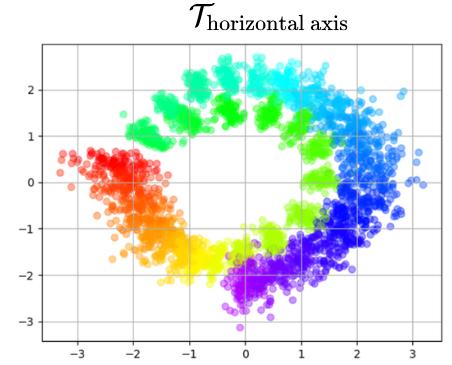}
  \caption{\small \textbf{Latent space visualizations on MPI3D.} Each 2D subspace $\mathcal{T}_i$ is plotted directly and colored by the factor it should encode.}
  \label{fig:latents_mpi}
\end{figure}

We plot the $T_i$ codes directly, using $\dim(\mathcal{T}_i)=2$ for visualization, and a 2D PCA projection of $S$. Each MPI3D factor is assigned to one $\mathcal{T}_i$ except one left free. In \cref{fig:latents_mpi}, every $T_i$ colored by its target factor exhibits clear structure. The free factor is also represented in $\mathcal{S}$ without explicit supervision, consistent with the organizing effect of \KL regularization \citep{burgess2018understandingdisentanglingbetavae}; that space appears less contrasted because the plot is a projection with 67\% variance explained. Full grids, with each latent space colored by each factor, are in Supp.~\ref{subsec:sup_latents}.

\subsection{Generations with attribute swapping}\label{sec:generations}
To visualize the latent interface, we swap factors between a \emph{source} and a \emph{target} image: we replace one $T_i$ component of the source by that of the target, keep every other latent component fixed, and decode. As shown in \cref{fig:teaser}, replacing a target block changes the corresponding attribute while largely preserving the rest of the representation; the same holds on dSprites, MPI3D and Cars3D (Supp.~\ref{sec:supp_generations}), orientation excepted, which is hard to read on some dSprites shapes. The backbone is a simple VAE prioritizing disentanglement over reconstruction quality, so the point is not photorealistic synthesis but direct intervention on a named latent block.

\subsection{Interpreting real-world biological variation on JUMP Cell Painting}
\label{sec:jump}
\begin{figure}[t]
  \centering
  \includegraphics[width=\linewidth]{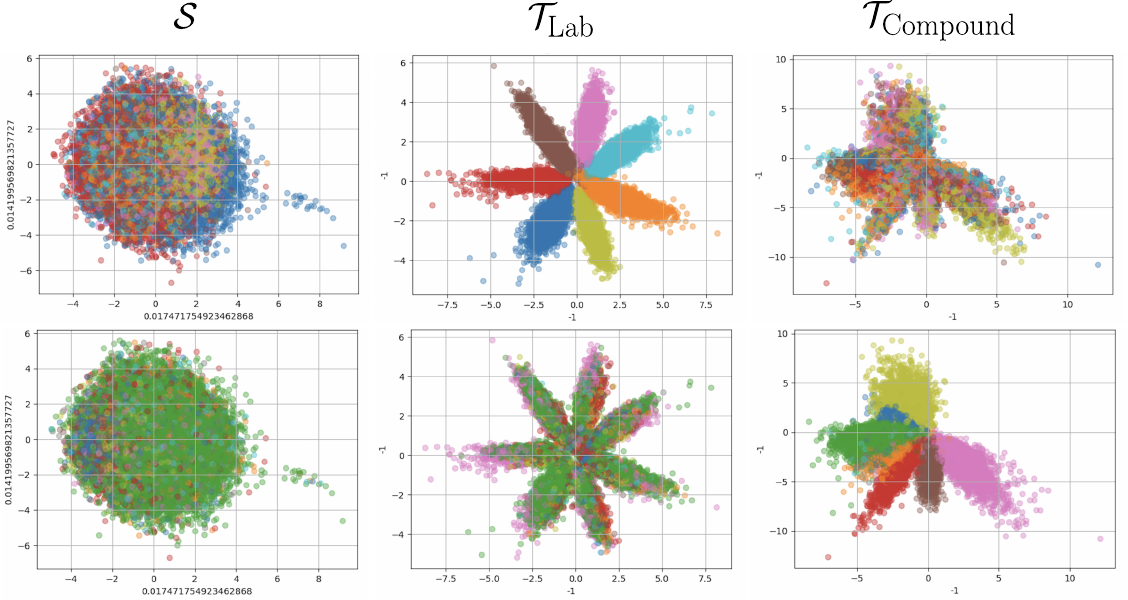}
  \caption{\small \textbf{Latent space visualizations on JUMP Cell Painting.} Compound perturbation and experimental source are encoded in distinct $\mathcal{T}$ subspaces, for 8 positive compounds and DMSO controls across 7 sources.}
  \label{fig:jump_latents}
\end{figure}

Beyond standard vision benchmarks, we consider the JUMP Cell Painting dataset \citep{chandrasekaran2023jump}. We select the 8 positive-control compounds and the DMSO negative control across 7 experimental sources, assign compound identity and source to two distinct $\mathcal{T}$ subspaces, and leave the remaining biological and technical variation in $\mathcal{S}$ (\cref{fig:jump_latents}). \textsc{XFactors} successfully disentangles the selected factors in this qualitative test on high-dimensional scientific images, suggesting that our method transfers to larger, more complex datasets with subtler factors.

\subsection{Scaling laws}
\label{sec:scaling_laws}
\begin{figure}[h]
  \centering
  \includegraphics[width=0.8\linewidth]{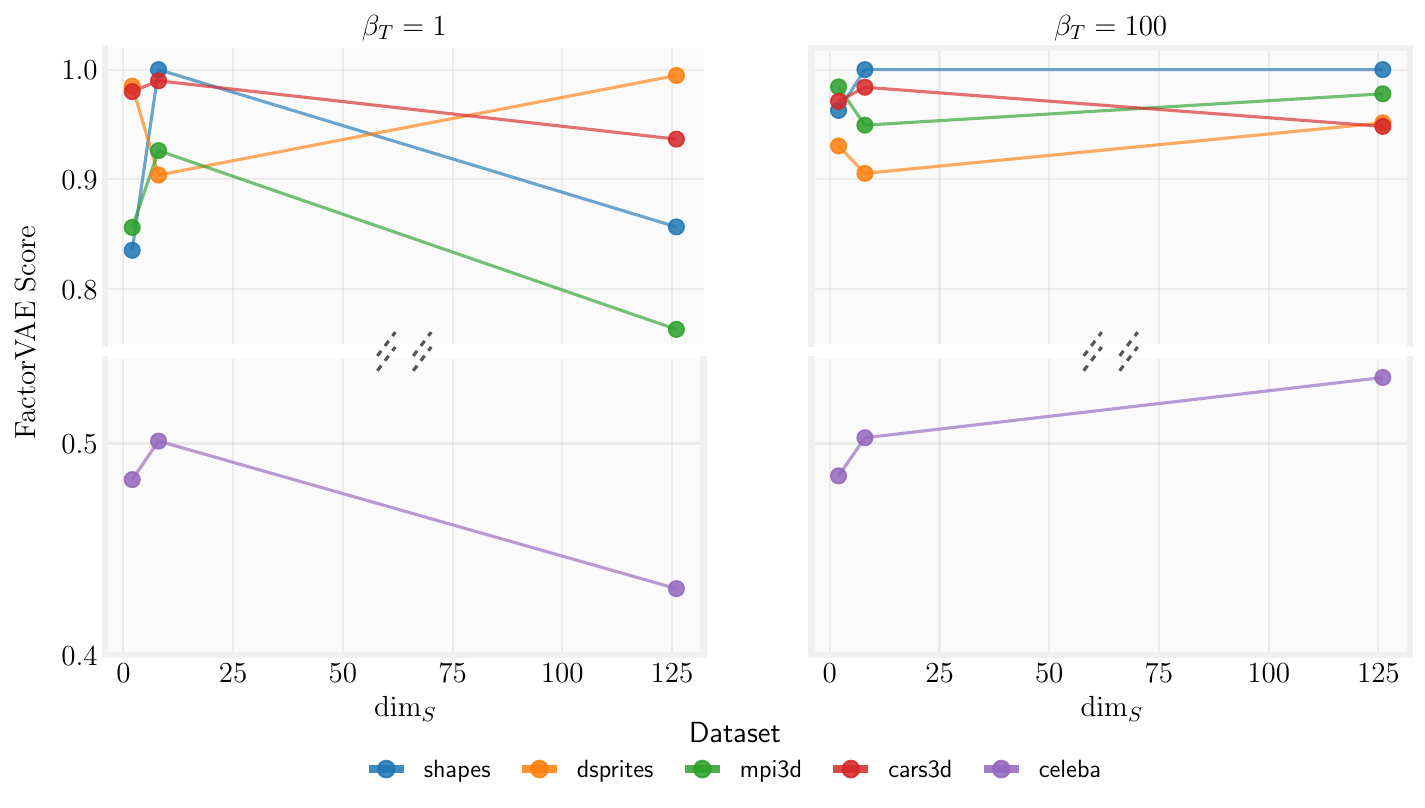}
  \caption{\small \textbf{Impact of residual dimension $\text{dim}_S$ and target regularization $\beta_T$.} Strong target regularization ($\beta_T=100$, right) prevents the decay in FactorVAE score that a growing $\text{dim}_S$ causes under weak regularization ($\beta_T=1$, left).}
  \label{fig:scaling_laws}
\end{figure}

We investigate the stability of \textsc{XFactors} with respect to the capacity of the residual subspace $\mathcal{S}$ and the regularization strength on the target subspace $\mathcal{T}$. \Cref{fig:scaling_laws} reports FactorVAE scores across five datasets as we vary the dimension of $\mathcal{S}$ ($\text{dim}_S$) and the target \KL weight $\beta_t$.

\paragraph{Scalability with $\text{dim}_S$.} A common failure mode in disentanglement methods is degraded performance when latent dimension exceeds the number of ground-truth factors. As shown in \cref{fig:scaling_laws}, increasing $\text{dim}_S$ up to 126 does not reduce the FactorVAE score when the target subspace is adequately regularized. This supports the use of a larger $\mathcal{S}$ to absorb uncontrolled or nuisance variation while keeping selected factors localized in $\mathcal{T}$.

\paragraph{Role of $\beta_t$.} The results further highlight an interaction between residual capacity and target regularization. Performance drops at higher dimensions under weak regularization ($\beta_t=1$, solid lines), while $\beta_t=100$ (dashed lines) mitigates this decay. Stronger regularization of $\mathcal{T}$ helps keep target factors localized as total latent capacity grows.

\subsection{Ablations and variations}
\label{sec:ablations}
\begin{table}[h]
  \centering
  \fontsize{8}{9.5}\selectfont
  \setlength{\tabcolsep}{3pt}
  \caption{\footnotesize \textsc{XFactors} vs variations and ablations on Shapes3D. Best value in \textbf{bold}, second best \underline{underlined}.}
  \label{tab:ablations}
  \begin{tabular}{c!{\color{gray!50}\vrule width 0.3pt}cc!{\color{gray!50}\vrule width 0.3pt}cc}
    \multirow{2}{*}{Metric} & \multicolumn{2}{c}{\textsc{XFactors}}              & \multicolumn{2}{c}{ablations}                                                                                             \\
                            & $\text{dim}_T=2$                                   & $\text{dim}_T=3$                                   & w/o InfoNCE                           & w/o $S$                      \\
    \midrule
    FactorVAE               & $\underline{\mathbf{1.000}}{\scriptstyle\pm .000}$ & $\underline{\mathbf{1.000}}{\scriptstyle\pm .000}$ & $.964{\scriptstyle\pm .046}$          & $.833{\scriptstyle\pm .000}$ \\
    D                       & $\underline{\mathbf{1.000}}{\scriptstyle\pm .000}$ & $\underline{\mathbf{1.000}}{\scriptstyle\pm .000}$ & $.933{\scriptstyle\pm .078}$          & $.829{\scriptstyle\pm .005}$ \\
    C                       & $\underline{.892}{\scriptstyle\pm .006}$           & $.855{\scriptstyle\pm .015}$                       & $\mathbf{.901}{\scriptstyle\pm .032}$ & $.684{\scriptstyle\pm .003}$ \\
    I                       & $\underline{.982}{\scriptstyle\pm .003}$           & $\mathbf{.984}{\scriptstyle\pm .011}$              & $.893{\scriptstyle\pm .084}$          & $.970{\scriptstyle\pm .021}$ \\
  \end{tabular}
\end{table}
We ablate \textsc{XFactors} on Shapes3D in \cref{tab:ablations} by removing the InfoNCE term, by removing $\mathcal{S}$, and by varying $\text{dim}_T$ from 2 to 3. Removing either $\mathcal{S}$ or InfoNCE degrades every metric: the contrastive signal is what localizes a factor in its block, and the residual space is what keeps unannotated variation out of the target blocks. Using a 3-dimensional $\mathcal{T}$ does not meaningfully change performance. See \ref{sec:annex-ablations} for further details.

\vspace{-2mm}\section{Discussion}\label{sec:discussion}\vspace{-2mm}
In this work, we present \textsc{XFactors}, a weakly supervised VAE framework for targeted disentanglement. Rather than attempting to recover all latent causes, we let the user select the labeled factors to expose. Each selected factor is assigned to a dedicated latent block $\mathcal{T}_i$, while the remaining variation is retained in a residual space $\mathcal{S}$ under the VAE bottleneck. We achieve this with a non-adversarial objective inspired by the Disentangled Information Bottleneck, using contrastive supervision to align each $\mathcal{T}_i$ with its corresponding factor and reconstruction/\KL regularization to preserve non-targeted information.

We evaluate \textsc{XFactors} on standard disentanglement benchmarks, CelebA, and JUMP Cell Painting. Across FactorVAE and DCI metrics, we observe strong performance compared with the literature, while factor-swapping experiments show that the learned blocks provide a direct handle for controlled generation. Our ablations confirm the role of the contrastive loss and the residual space, and our scaling experiments show that increasing the capacity of $\mathcal{S}$ does not necessarily degrade the target-factor interface when the target space is sufficiently regularized.

The main limitation of the current model is reconstruction quality, especially on CelebA, yet not affecting the disentanglement objective adressed. This is partly due to the classical VAE trade-off between reconstruction fidelity and \KL regularization, and suggests combining our factor interface with stronger generative backbones. We also believe that partial disentanglement requires more systematic evaluation. Existing benchmarks often blur the distinction between fully factorized datasets with all factor combinations available and realistic datasets, such as CelebA, where attributes are correlated and coverage is incomplete. In order for such an interface to be trusted for discovery, future work should therefore include broader leakage diagnostics between $\mathcal{S}$ and the $\mathcal{T}_i$ blocks, correlation-shift tests, and held-out-combination evaluations.

Finally, we hope that the JUMP results pave the way to novel biological discoveries, from batch-effect inspection and removal to generating testable counterfactuals, enabling for example important speedups in drug discovery pipelines.

\clearpage

\bibliography{bib}
\bibliographystyle{plainnat}

\clearpage

\appendix

\section{Supplementary Materials}\label{sec:supp}

\subsection{Societal impacts}
\label{sec:societal_impacts}
Through counterfactual generation, \textsc{XFactors} makes it possible to edit the visual factors of an image. To date, the risk of generating counterfactuals of human faces such as CelebA is limited by the reconstruction quality of the VAE. On less complex images, however, counterfactual generation could prove more convincing.

\subsection{Related Work}\label{sec:related_work}

\paragraph{Information Theory \& Disentanglement.} The Information Bottleneck (IB) principle \citep{tishby2000informationbottleneckmethod} formalizes representation learning as a trade-off between compressing the input and retaining task-relevant information, with variational deep extensions proposed by \citet{alemi2016deepvariationalinformationbottleneck}. DisenIB \citep{pan2020disentangledinformationbottleneck} further partitions the latent space into task-relevant and nuisance components, but typically relies on adversarial or min-max objectives to control mutual information terms. \textsc{XFactors} keeps the structural split while replacing adversarial games with explicit latent blocks, reconstruction, KL regularization, and contrastive supervision.

\paragraph{Supervision and Inductive Biases.} Prior work shows that fully unsupervised disentanglement is not identifiable without inductive biases \citep{locatello2018challengingcommonassumptionsunsupervised}, and that learned axes may follow dataset correlations rather than user-relevant semantics \citep{träuble2021disentangledrepresentationslearnedcorrelated}. Labels can resolve part of this ambiguity by guiding factorization \citep{mathieu2016disentanglingfactorsvariationdeep, lample2018fadernetworksmanipulatingimages}, but complex datasets seldom annotate every source of variation. This motivates the study of a partial setting: only targeted factors are labeled, while residual variation is preserved for downstream use.

\paragraph{Contrastive and Generative Approaches.} Contrastive learning supplies a direct way to shape latent geometry: InfoNCE \citep{oord2019representationlearningcontrastivepredictive} pulls positive pairs together and separates negatives, and has been used as a regularizer for latent discovery methods such as DisCo \citep{ren2021learningdisentangledrepresentationexploiting}. We use it as the main supervised signal for each selected factor block. In generative modeling, $\beta$-VAE \citep{higgins2017betavae, burgess2018understandingdisentanglingbetavae} promotes factorization through stronger KL regularization but faces the usual rate-distortion trade-off. GAN-based methods such as InfoGAN-CR~\citep{lin2020infogancrmodelcentralityselfsupervisedmodel}, GAN-LD~\citep{voynov2020unsuperviseddiscoveryinterpretabledirections}, DisCo, and GANSpace~\citep{härkönen2020ganspacediscoveringinterpretablegan} can reveal useful directions, but often require adversarial training or post-hoc latent analysis. Diffusion approaches such as DisDiff \citep{yang2023disdiffunsuperviseddisentanglementdiffusion} and EncDiff \citep{yang2024diffusionmodelcrossattention} offer powerful generation, yet editing may rely on iterative sampling, concept mechanisms, or dataset-specific tuning.

\paragraph{Targeted Factor Interfaces.} Attribute-supervised editing methods such as Fader Networks use labels to manipulate selected factors through conditional generation \citep{lample2018fadernetworksmanipulatingimages}. Closer to our setting, supervised adversarial disentangling and DisenIB separate task-relevant information from complementary nuisance variation \citep{mathieu2016disentanglingfactorsvariationdeep, pan2020disentangledinformationbottleneck}, while SCBD learns block-structured embeddings for target and environment-dependent factors \citep{makino2025supervisedcontrastiveblockdisentanglement}. \textsc{XFactors} differs by assigning a separate latent block to each selected factor while retaining an explicit residual VAE subspace for reconstruction and inspection.

To this end, our approach avoids the pitfalls of classifier-based disentanglement, known to be prone to adversarial perturbations \citep{szegedy2014intriguingpropertiesneuralnetworks, augustin2022diffusionvisualcounterfactualexplanations}, and instead relies on a robust contrastive learning objective. It also prevents the often overlooked interpretability problem of unsupervised methods, which necessitate a tedious manual search over their obtained representation spaces in order to link each one to a true semantic factor \citep{locatello2018challengingcommonassumptionsunsupervised}.

\subsection{Theoretical foundations}

\paragraph{The Information Bottleneck.}
Let $p$ be a probability distribution on $\mathbb{R}^m$; $p$ is \emph{disentangled} if it can be factorized as:
\begin{equation}
    \begin{split}
        p(x_1,x_2,\ldots,x_K) = \prod_{i=1}^K p_i(\mathbf{x}_i) \quad\text{s.t.}\quad \mathbf{x}_i\in\mathbb{R}^{m_i} \quad\text{and}\quad \sum_{i=1}^K m_i = m
    \end{split}
    \label{eq:factorized_distribution}
\end{equation}
The objective in \cref{eq:factorized_distribution} is to obtain $K$ mutually independent marginal distributions on $\mathbb{R}^{m_i}$. Each marginal distribution may then capture a specific feature of the data. In this paper, we refer to such a feature as a \emph{factor of variation}, or simply a \textbf{factor}. Accordingly, the goal of disentangling a latent space is to learn a latent representation whose distribution factorizes into independent components, such that each marginal distribution corresponds to one factor.

Let $X \subset \mathcal{X}$ denote our dataset, with $\mathcal{X}$ the pixel space and let $Y \subset \mathcal{Y}$ denote the associated factors. We denote $Z \subset \mathcal{Z}$ the set of representations obtained from $X$, with $\mathcal{Z}$ their embedding space.

The Information Bottleneck principle~\cite{tishby2000informationbottleneckmethod} proposes to map $X$ to a latent representation $Z$ in a latent space $\mathcal{Z}$ by balancing two competing objectives: (i) compressing the information that $Z$ retains about $X$, and (ii) preserving the information that $Z$ carries about $Y$.

This trade-off is classically formulated by optimizing the following Lagrangian:
\begin{equation}
    \mathcal{L}_{\mathrm{IB}}(\beta) = \beta\,\mathcal{I}(X;Z) - \mathcal{I}(Z;Y)
    \label{eq:lagrang_ib}
\end{equation}
where $\mathcal{I}(\cdot\,;\cdot)$ denotes the Shannon mutual information (MI).

\citet{burgess2018understandingdisentanglingbetavae} connect the Information Bottleneck Lagrangian in \cref{eq:lagrang_ib} to the $\beta$-VAE objective:
\begin{equation}
    \mathcal{L}_{\beta\mathrm{VAE}}
    = \mathbb{E}_{q_\phi(z\mid x)}\big[\log p_\theta(x\mid z)\big] - \beta\,\mathrm{KL}\big(q_\phi(z\mid x)\,\|\,p(z)\big)
    \label{eq:beta_vae}
\end{equation}
Their experiments report \emph{natural} disentanglement that typically increases with $\beta$ in \cref{eq:beta_vae}. However, this disentanglement remains implicit: the model does not expose \emph{which} factors are encoded, nor \emph{where}. They also relate the $\mathrm{KL}$ term to Shannon mutual information, linking $\beta$-VAE training to the IB principle.

This information-theoretic framework provides a practical formulation for \emph{supervised} disentanglement, as proposed by \citet{pan2020disentangledinformationbottleneck}. They partition the latent space $\mathcal{Z}$ into two subspaces, $\mathcal{S}$ and $\mathcal{T}$, in order to induce a factorization of the latent distribution,
\begin{equation}
    q_\theta(\mathbf{z}) = q_{\theta_1}(\mathbf{z}_s)\, q_{\theta_2}(\mathbf{z}_t)
    \label{eq:disenib_factorization}
\end{equation}
where $\mathbf{z}_s \in \mathcal{S}$ and $\mathbf{z}_t \in \mathcal{T}$. In \cref{eq:disenib_factorization}, $\mathcal{T}$ is encouraged to encode a selected labeled factor $\mathit{y_f}$, whereas $\mathcal{S}$ is intended to absorb other variation.

Following \citet{pan2020disentangledinformationbottleneck}, the Information Bottleneck objective in \cref{eq:lagrang_ib} is reformulated, with $S \subset \mathcal{S}$ and  $T \subset \mathcal{T}$, as
\begin{equation}
    \mathcal{L}_{\mathrm{DisenIB}} = -\mathcal{I}(T;Y) - \mathcal{I}\big(X; (S,Y)\big)+ \mathcal{I}\big(S;T\big)
    \label{eq:disenib}
\end{equation}

They approximate these mutual information terms using auto-encoder networks trained with adversarial and classification objectives, focusing on the split between one selected factor and the rest.

\paragraph{Contrastive Supervision.}
Disentanglement is closely tied to latent geometry. Contrastive learning provides a tractable signal for promoting factor alignment by shaping each target block according to factor-based similarity. The InfoNCE objective introduced by \citet{oord2019representationlearningcontrastivepredictive} pulls latent vectors with the same factor value together and pushes vectors with different values apart. Its connection to mutual-information estimation \citep{oord2019representationlearningcontrastivepredictive, poole2019variationaleboundsmi} motivates its use here, but we use it as a practical contrastive surrogate rather than an exact MI maximization procedure.

Concretely, for factor $i$ and a batch $\mathcal{B}$, let $\mathbf{t}^{(i)}_a$ be the code of anchor $a$ in block $\mathcal{T}_i$ and $P_i(a)=\{p\in\mathcal{B}\setminus\{a\}:y_{f_i}^p=y_{f_i}^a\}$ its positives. With similarity $\operatorname{sim}$ and temperature $\tau$, we use the supervised batch loss:
\begin{equation}
    \mathcal{L}_{\mathrm{InfoNCE}}^{(i)}
    =
    -\frac{1}{|\mathcal{B}|}\sum_{a\in\mathcal{B}}
    \frac{1}{|P_i(a)|}
    \sum_{p\in P_i(a)}
    \log
    \frac{\exp(\operatorname{sim}(\mathbf{t}^{(i)}_a,\mathbf{t}^{(i)}_p)/\tau)}
    {\sum_{b\in\mathcal{B}\setminus\{a\}}\exp(\operatorname{sim}(\mathbf{t}^{(i)}_a,\mathbf{t}^{(i)}_b)/\tau)}
    \label{eq:supervised_infonce}
\end{equation}
Anchors with no positive in the batch are ignored for that factor.

Unlike adversarial MI estimators such as MINE \citep{belghazi18mine}, \cref{eq:supervised_infonce} is a direct minimization objective. For the usual finite-candidate setting, InfoNCE also yields the motivational lower-bound relation:

\begin{equation}
    I(T_i;y_{f_i}) \ge \log N - \mathcal{L}_{\mathrm{InfoNCE}}, \quad N = \mid T_i \mid
    \label{eq:infonce_bound}
\end{equation}

\subsection{Algorithms}
\paragraph{Concrete algorithms}
The algorithm for training and inference are given in \cref{alg:xfactors-concise} and \cref{alg:xfactors-inference-transfer}.

\begin{figure}[t]
    \centering
    \begin{minipage}[t]{0.48\linewidth}
            \vspace{0pt}
            \begin{minipage}[t][2\baselineskip][t]{\linewidth}
            \captionof{algorithm}{\textsc{XFactors} Training}
            \label{alg:xfactors-concise}
            \end{minipage}
            \small
            \begin{algorithmic}[1]
                \Require Images $x$, factors $y_{f_i}$, encoder $\psi_s(\cdot)$, encoder $\psi_t(\cdot)$, decoder $\phi(\cdot)$
                \For{each $(x,y)$ in batch}
                \State $(\mu_s,\sigma_s),(\mu_t,\sigma_t) \leftarrow \psi_s(x), \psi_t(x)$
                \State $z_s,z_t \leftarrow \mathcal{N}(\mu_s,\sigma_s), \mathcal{N}(\mu_t,\sigma_t)$
                \State $\hat{x} \leftarrow \phi(z_s\|z_t)$
                \State $\mathcal{L}_{\text{reco}} \leftarrow \text{MSE}(\hat{x},x)$
                \State $\mathcal{L}_{\text{KL}} \leftarrow (\beta_s\text{KL}_s+\beta_t\text{KL}_t)$
                \State $\mathcal{L}_{\text{NCE}} \leftarrow \sum_i \lambda_i \cdot \text{InfoNCE}(z_t^{(i)}, y_{{f_i}})$
                \State $\mathcal{L} \leftarrow$ $\mathcal{L}_{\text{reco}}$ + $\mathcal{L}_{\text{KL}}$ + $\mathcal{L}_{\text{NCE}}$
                \State Update parameters using $\nabla\mathcal{L}$
                \EndFor
            \end{algorithmic}
    \end{minipage}
    \hfill
    \begin{minipage}[t]{0.48\linewidth}
            \vspace{0pt}
            \begin{minipage}[t][2\baselineskip][t]{\linewidth}
            \captionof{algorithm}{\textsc{XFactors} Inference\\{\small(source-target attribute swapping case)}}
            \label{alg:xfactors-inference-transfer}
            \end{minipage}
            \small
            \begin{algorithmic}[1]
                \Require Source image $x^\text{src}$, target image $x^\text{tgt}$, desired factor ${f}_i$, encoders $\psi_s,\psi_t$, decoder $\phi$
                \State Encode source: $(\mu_s^\text{src},\_),(\mu_t^\text{src},\_) \leftarrow \psi_s(x^\text{src}), \psi_t(x^\text{src})$
                \State Encode target: $(\mu_s^\text{tgt},\_),(\mu_t^\text{tgt},\_) \leftarrow \psi_s(x^\text{tgt}), \psi_t(x^\text{tgt})$
                \State Use $z_s$ from source: $z_s \leftarrow \mu_s^\text{src}$
                \State Combine factors from source and target into $z_t$:\\$z_t \leftarrow (\mu_{t,f_1}^\text{src}, \ldots, \mu_{t,f_i}^\text{tgt}, \ldots, \mu_{t,f_K}^\text{src})$
                \State Decode: $x_{\text{edit}} \leftarrow \phi(z_s\|z_t)$
            \end{algorithmic}
    \end{minipage}
\end{figure}

\subsection{Architecture \hyperref[sec:archi]{$\upuparrows$}}
The overall architecture is depicted in \cref{fig:archi} of the main text.
Our method does not imply a precise architecture for the encoder/decoder part.
As a proof of concept, and since the generation/reconstruction are not our main tasks, we choose the architecture:

\begin{table}[h]
    \centering
    \footnotesize
    \setlength{\tabcolsep}{4pt}
    \renewcommand{\arraystretch}{1.05}
    \caption{\textbf{Encoder.} Conv$3\times3$ with stride 2 halves spatial resolution (up to ceiling). \texttt{ResBlock} is \texttt{SimpleConv} (Conv$3\times3$--BN--LeakyReLU) or \texttt{ResidualBlock}. VAE head outputs $(\boldsymbol{\mu},\log\boldsymbol{\sigma}^2)\in\mathbb{R}^d\times\mathbb{R}^d$.}
    \label{tab:encoder_arch_compact}
    \begin{tabular}{lll}
        \toprule
        \textbf{Layer}        & \textbf{Channels}          & \textbf{Spatial}                         \\
        \midrule
        Input                 & $C_{\mathrm{in}}$          & $H\times W$                              \\
        Conv$3$/s2 + BN + Act & $C_{\mathrm{in}}\!\to\!48$ & $\lceil H/2\rceil\times\lceil W/2\rceil$ \\
        ResBlock              & $48\!\to\!48$              & $\lceil H/2\rceil\times\lceil W/2\rceil$ \\
        Conv$3$/s2 + BN + Act & $48\!\to\!96$              & $\lceil H/4\rceil\times\lceil W/4\rceil$ \\
        ResBlock              & $96\!\to\!96$              & $\lceil H/4\rceil\times\lceil W/4\rceil$ \\
        Conv$3$/s2 + BN + Act & $96\!\to\!192$             & $\lceil H/8\rceil\times\lceil W/8\rceil$ \\
        Conv$3$/s1 + Act      & $192\!\to\!192$            & $\lceil H/8\rceil\times\lceil W/8\rceil$ \\
        Flatten + Linear      & $\to 2d$                   & --                                       \\
        \bottomrule
    \end{tabular}
\end{table}

\begin{table}[h]
    \centering
    \footnotesize
    \setlength{\tabcolsep}{4pt}
    \renewcommand{\arraystretch}{1.05}
    \caption{\textbf{Decoder.} Conv $4\times4$ (stride 2, pad 1) doubles spatial resolution.}
    \label{tab:decoder_arch_compact}
    \begin{tabular}{lll}
        \toprule
        \textbf{Layer}         & \textbf{Channels}                         & \textbf{Spatial}  \\
        \midrule
        Input latent           & $d$                                       & --                \\
        Linear + Unflatten     & $\to C_e$                                 & $H_e\times W_e$   \\
        Conv$3$/s1 + BN + Act  & $192\!\to\!192$                           & $H_e\times W_e$   \\
        ConvT$4$/s2 + BN + Act & $192\!\to\!96$                            & $2H_e\times 2W_e$ \\
        ResBlock               & $96\!\to\!96$                             & $2H_e\times 2W_e$ \\
        ConvT$4$/s2 + BN + Act & $96\!\to\!48$                             & $4H_e\times 4W_e$ \\
        ResBlock               & $48\!\to\!48$                             & $4H_e\times 4W_e$ \\
        ConvT$4$/s2 + Act      & $48\!\to\!C_{\mathrm{out}}$               & $8H_e\times 8W_e$ \\
        Conv$3$/s1             & $C_{\mathrm{out}}\!\to\!C_{\mathrm{out}}$ & $8H_e\times 8W_e$ \\
        Sigmoid                & --                                        & $H\times W$       \\
        \bottomrule
    \end{tabular}
\end{table}

\subsection{Experiments \hyperref[sec:experiments]{$\upuparrows$}}
We evaluate \textsc{XFactors} on standard disentanglement benchmarks, CelebA, and JUMP Cell Painting. Across FactorVAE and DCI metrics, we observe strong performance compared with the literature, while factor-swapping experiments show that the learned blocks provide a direct handle for controlled generation. Our ablations confirm the role of the contrastive loss and the residual space, and our scaling experiments show that increasing the capacity of $\mathcal{S}$ does not necessarily degrade the target-factor interface when the target space is sufficiently regularized.

We used the same set of \textsc{XFactors} hyperparameters for all datasets: $\text{dim}_{\mathcal{T}_i} = 2$, $\text{dim}_\mathcal{S} = 126$, $\beta_t = 100$, $\beta_s = 100$.

We train XFactors on an NVIDIA A100 in 15 hours for 3DShapes.

\subsection{Disentanglement metrics \hyperref[sec:metrics]{$\upuparrows$}}
\label{sec:supp_disentanglement_metrics_details}
The methods reported in \cref{tab:disentanglement-metrics} are: FactorVAE~\citep{kim2019disentanglingfactorising}, $\beta$-TCVAE~\citep{chen2018isolatingsourcesdisentanglementvariational}, EncDiff~\citep{yang2024diffusionmodelcrossattention}, DSD~\citep{feng2018dualswapdisentangling}, Ada-ML-VAE and Ada-GVAE~\citep{locatello20a}, and SW-VAE~\citep{zhu2022swvae}.\\
Metrics of unsupervised methods reported in \cref{tab:disentanglement-metrics} come from \citet{yang2024diffusionmodelcrossattention}, which seems to obtain them from \citet{ren2021learningdisentangledrepresentationexploiting}. dSprites values for $\beta$-TCVAE are from their respective papers \citep{chen2018isolatingsourcesdisentanglementvariational}. We took Ada-GVAE and Ada-ML-VAE metrics directly from \citet{locatello20a}, and DSD and SW-VAE metrics from \citet{zhu2022swvae}.

\begin{table}[h]
  \centering
  \fontsize{6}{8.5}\selectfont
  \setlength{\tabcolsep}{3pt}
  \caption{\footnotesize CelebA disentanglement metrics for $n$ disentangled attributes, for weakly \underline{S}upervised and \underline{U}nsupervised methods. Sources of the reported numbers are given below; the full DCI triplet for \textsc{XFactors} is in Table~\ref{tab:supp_xfactors_dci_triplets}.}
  \label{tab:celeba_metrics}
  \resizebox{0.62\linewidth}{!}{%
  \begin{tabular}{c|c|c|cc}
                                                                 & $n$               & Method                        & FactorVAE                     & DCI                           \\
    \midrule \multirow{2}{*}{S}                                  &
    5                                                            & \textsc{XFactors} & $.532{\scriptstyle \pm .002}$ & $.690{\scriptstyle \pm .001}$                                 \\
    \arrayrulecolor{gray!50}\cmidrule{2-5}\arrayrulecolor{black} &
    2                                                            & CMI               & $1.0{\scriptstyle \pm .0}$    & $.807{\scriptstyle \pm .023}$                                 \\
    \hline \multirow{3}{*}{U}                                    &
    \multirow{3}{*}{$40$}                                        & FactorVAE         & $.120{\scriptstyle \pm .021}$ & $.071{\scriptstyle \pm .007}$                                 \\
                                                                 &                   & $\beta$-TCVAE                 & $.098{\scriptstyle \pm .024}$ & $.035{\scriptstyle \pm .011}$ \\
                                                                 &                   & InfoGAN-CR                    & $.113$                        & $.220$                        \\
  \end{tabular}
  }
\end{table}

For \cref{tab:celeba_metrics}, metrics for FactorVAE , $\beta$-TCVAE and InfoGAN-CR come from \citet{NEURIPS2022_0850e04a} where the 40 attributes are tentatively disentangled.
We (\textsc{XFactors}) disentangle 5 factors of CelebA and leave the rest in $\mathcal{S}$. CMI \citep{FunkeVWKZB22} disentangles only 2 attributes.

We report standard deviations w.r.t. the computation of metrics for \textsc{XFactors} (we run 5 evaluations).

\paragraph{Computational details for FactorVAE-score}
Following \citet{kim2019disentanglingfactorising}, we compute FactorVAE-score with the following algorithm:

\begin{algorithm}
    \caption{FactorVAE Computation}\label{alg:factorvae}
    \begin{algorithmic}[1]
        \State \textbf{Inputs:} Factors $(f_1,\ldots,f_K)$, standardized latent codes $\{ \mathbf{z}_i \}$, number of iterations $n_{\mathrm{iter}}$.
        \For{$n = 1$ \textbf{to} $n_{\mathrm{iter}}$}
        \State Sample $k \sim \mathcal{U}\{1,\ldots,K\}$
        \State Select the factor $f_k$ and sample a factor value $v_k$
        \State Select the subset of examples such that $f_k = v_k$
        \State Compute the per-dimension variances $\sigma_d^2 = \operatorname{Var}(z_{i,d})$ for all $d$
        \State Identify $d^* = \arg\min_d \sigma_d^2$
        \State Store the pair $(d^*, f_k)$
        \EndFor
        \State Train a majority-vote classifier on the collected pairs $(d^*, f_k)$
        \State \textbf{Output:} $\mathrm{score}_{\mathrm{FactorVAE}} = \mathrm{Acc}(f_k \mid d^*)$
    \end{algorithmic}
\end{algorithm}

Hyperparameters for the \cref{alg:factorvae} are the batch size of the subset at \textit{line 5} and $n_\mathrm{iter}$.

We choose to run:
\begin{table}[h]
    \centering
    \caption{Hyperparameters used to compute the FactorVAE}
    \label{tab:factorvae_hparams}
    \begin{tabular}{ll}
        \toprule
        \textbf{Hyperparameter} & \textbf{Value} \\
        \midrule
        $n_{\mathrm{iter}}$     & $10^4$         \\
        $\mathrm{batch\_size}$  & $64$           \\
        \bottomrule
    \end{tabular}
\end{table}

\begin{figure}
    \centering
    \includegraphics[width=0.7\linewidth]{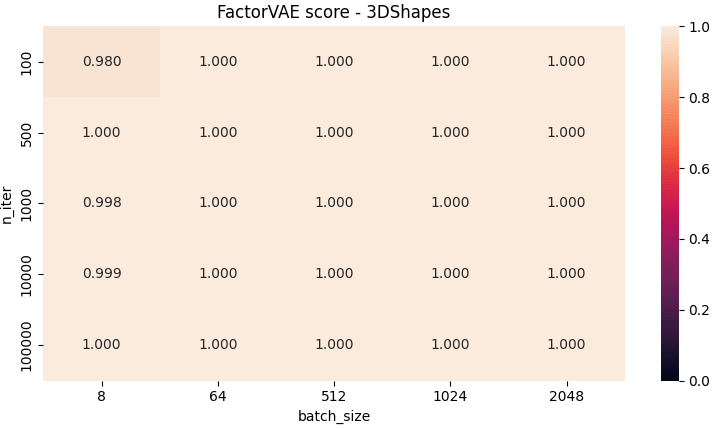}
    \caption{FactorVAE scores on Shapes3D for different batch sizes and numbers of iterations.}
    \label{fig:heatmap_shapes}
\end{figure}

However our metrics remain stable across the variation of these hyperparameters as can be seen in \cref{fig:heatmap_shapes}.

\paragraph{Other implementation for FactorVAE-score}
EncDiff \cite{yang2024diffusionmodelcrossattention}, DisCo \cite{ren2021learningdisentangledrepresentationexploiting} and DisDiff \cite{yang2023disdiffunsuperviseddisentanglementdiffusion}, choose to fit a PCA at the first order on their subspaces before computing the FactorVAE-score.
We choose to stay close to the \citet{kim2019disentanglingfactorising} algorithm.

\paragraph{DCI computational details}
Following \citet{DCI} we compute DCI metric as a triplet.

We choose as a classifier a Random Forest Classifier.

\begin{table}[h]
    \centering
    \caption{Hyperparameters used to compute the DCI metrics.}
    \label{tab:dci_hparams}
    \begin{tabular}{ll}
        \toprule
        \textbf{Hyperparameter}  & \textbf{Value} \\
        \midrule
        $\mathrm{n\_estimators}$ & $20$           \\
        $\mathrm{max\_depth}$    & $20$           \\
        \bottomrule
    \end{tabular}
\end{table}

\paragraph{Full DCI triplets for \textsc{XFactors}}
Because most baselines only report the disentanglement component $D$, we report the full DCI triplets here for completeness.

\begin{table}[h]
    \centering
    \caption{Full DCI triplets for \textsc{XFactors}.}
    \label{tab:supp_xfactors_dci_triplets}
    \begin{tabular}{lccc}
        \toprule
        \textbf{Dataset} & \textbf{D}                     & \textbf{C}                    & \textbf{I}                     \\
        \midrule
        Shapes3D         & $1.000{\scriptstyle \pm .000}$ & $.885{\scriptstyle \pm .001}$ & $1.000{\scriptstyle \pm .000}$ \\
        MPI3D            & $.949{\scriptstyle \pm .000}$  & $.797{\scriptstyle \pm .000}$ & $.997{\scriptstyle \pm .000}$  \\
        dSprites         & $.909{\scriptstyle \pm .001}$  & $.665{\scriptstyle \pm .001}$ & $.858{\scriptstyle \pm .000}$  \\
        cars3D           & $.802{\scriptstyle \pm .004}$  & $.505{\scriptstyle \pm .002}$ & $.695{\scriptstyle \pm .005}$  \\
        CelebA           & $.690{\scriptstyle \pm .001}$  & $.626{\scriptstyle \pm .001}$ & $.814{\scriptstyle \pm .001}$  \\
        \bottomrule
    \end{tabular}
\end{table}

These metrics are most informative when interpreted at the level of the targeted interface. On the synthetic datasets, most annotated generative factors are available and comparisons are closer to the classical full-disentanglement setting. On CelebA, by contrast, \textsc{XFactors} intentionally supervises only five attributes and leaves the remaining facial variation in $\mathcal{S}$; the reported scores therefore assess whether the selected attributes are localized, not whether all facial causes have been recovered.

\paragraph{Computing FactorVAE-score and DCI with partial disentanglement}
XFactors can perform disentanglement even with a small subset of the factors thanks to the decomposition of the latent spaces in \cref{eq:directsum}.

\paragraph{CelebA details}
We assess the performance when multiple factors remain in $\mathcal{S}$ with CelebA, where 5 factors were disentangled in $\bigoplus_{i=1}^5 \mathcal{T}_i$ and other factors were in $\mathcal{S}$.

Since we do not try to disentangle the 35 factors in $\mathcal{S}$ but only to disentangle them, as a block, from the 5 others in $\bigoplus_{i=1}^5 \mathcal{T}_i$, we cannot compute FactorVAE-score or DCI with all the factors.

We consider all the factors in $\mathcal{S}$ as one factor \texttt{"s"}.

\subsection{Ablations and variations \hyperref[sec:experiments]{$\upuparrows$}}
\label{sec:annex-ablations}

\begin{figure*}[h]
    \centering
    \includegraphics[width=\linewidth]{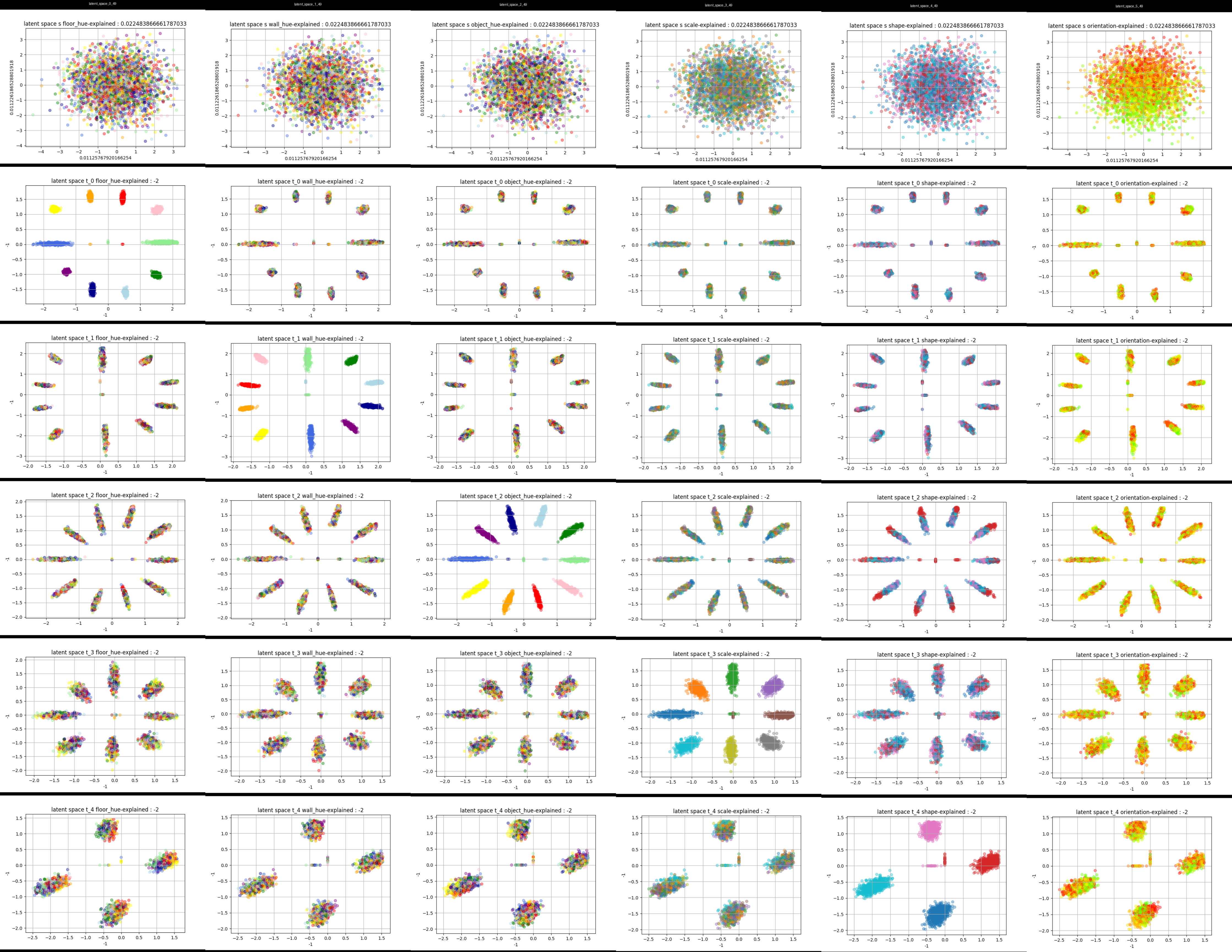}
    \caption{\footnotesize \textbf{Learned latents on Shapes3D}. Rows: encoding space (top to bottom: $(\mathcal{S}, \mathcal{T}_0, \mathcal{T}_1, \mathcal{T}_2, \mathcal{T}_3, \mathcal{T}_4)$). Columns: color being used, corresponding to the different possible values of a single factor (left to right: (floor hue, wall hue, object hue, scale, shape, orientation), only the coloring changes between columns). For the $\mathcal{S}$ space the first 2 components of its PCA are shown.}
    \label{fig:supp_no_ablation}
\end{figure*}

\paragraph{Baseline}
We present in \cref{fig:supp_no_ablation} the latents obtained by XFactors –without any modification– on the validation set of Shapes3D. For each factor $i$, the related encoding $T_i$ is well-structured, while other representation spaces appear largely unstructured, indicating strong block-specific organization.

\begin{figure*}[h]
    \centering
    \includegraphics[width=\linewidth]{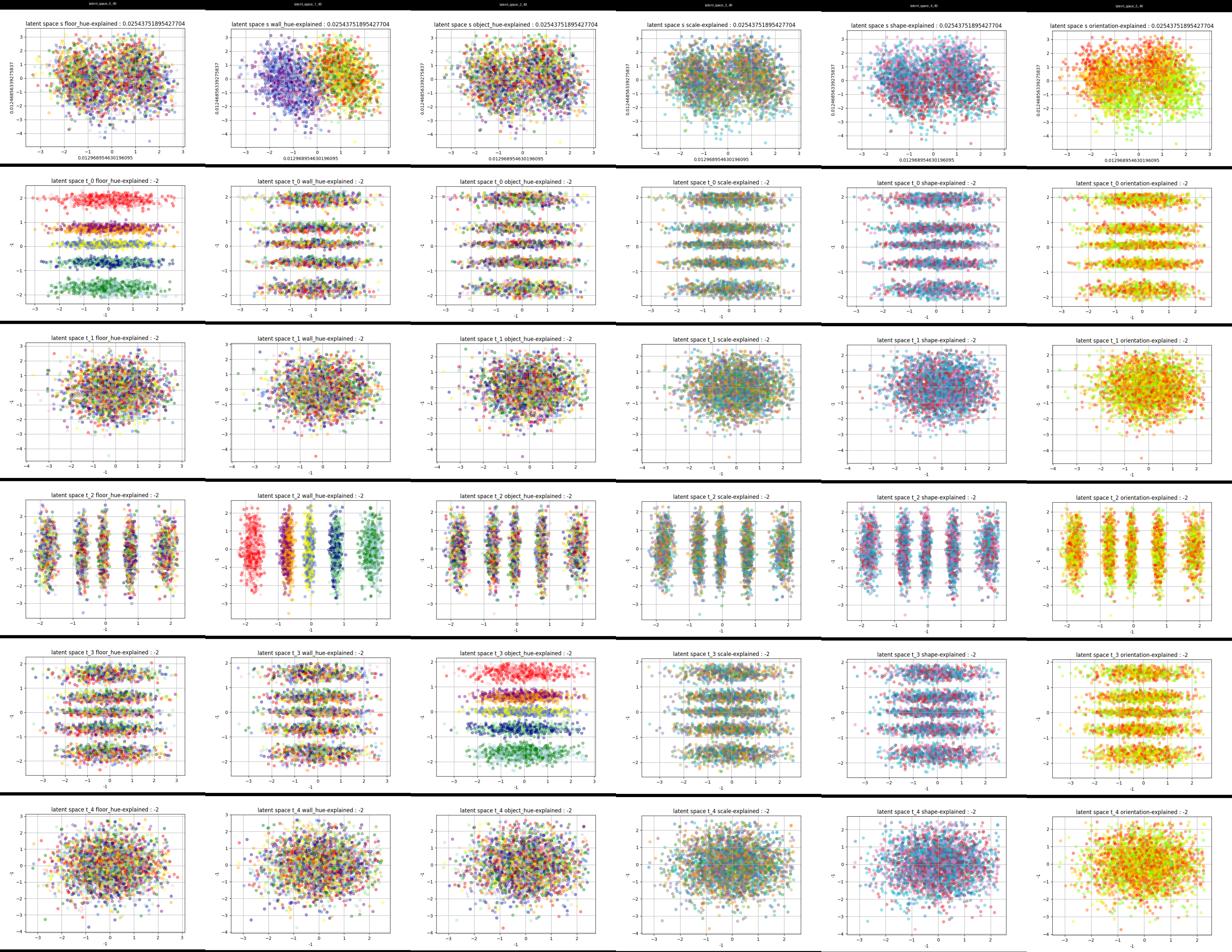}
    \caption{\footnotesize \textbf{Learned latents on Shapes3D when the InfoNCE term of the loss is ablated}. Rows: encoding space (top to bottom: $(\mathcal{S}, \mathcal{T}_0, \mathcal{T}_1, \mathcal{T}_2, \mathcal{T}_3, \mathcal{T}_4)$). Columns: color being used, corresponding to the different possible values of a single factor (left to right: (floor hue, wall hue, object hue, scale, shape, orientation), only the coloring changes between columns). For the $\mathcal{S}$ space the first 2 components of its PCA are shown.}
    \label{fig:supp_no_info_nce}
\end{figure*}

\paragraph{No InfoNCE}
Compare them to those obtained by ablating the InfoNCE term of our loss, in \cref{fig:supp_no_info_nce}. We can observe 2 behaviors: firstly, some $T$ encodings are totally entangled (\textit{e.g.} $T_1$ at 3rd row), and conversely some factors are totally entangled (\textit{e.g.} shape at 5th column), indicating only partial disentanglement might have been achieved on some dimensions. Secondly, even the seemingly structured $T$ encodings are much less well structured as important overlaps can be seen between clusters of different values for some factors (\textit{e.g.} (1st column, 2nd row), or (3rd column, 5th row)).

\begin{figure*}[h]
    \centering
    \includegraphics[width=\linewidth]{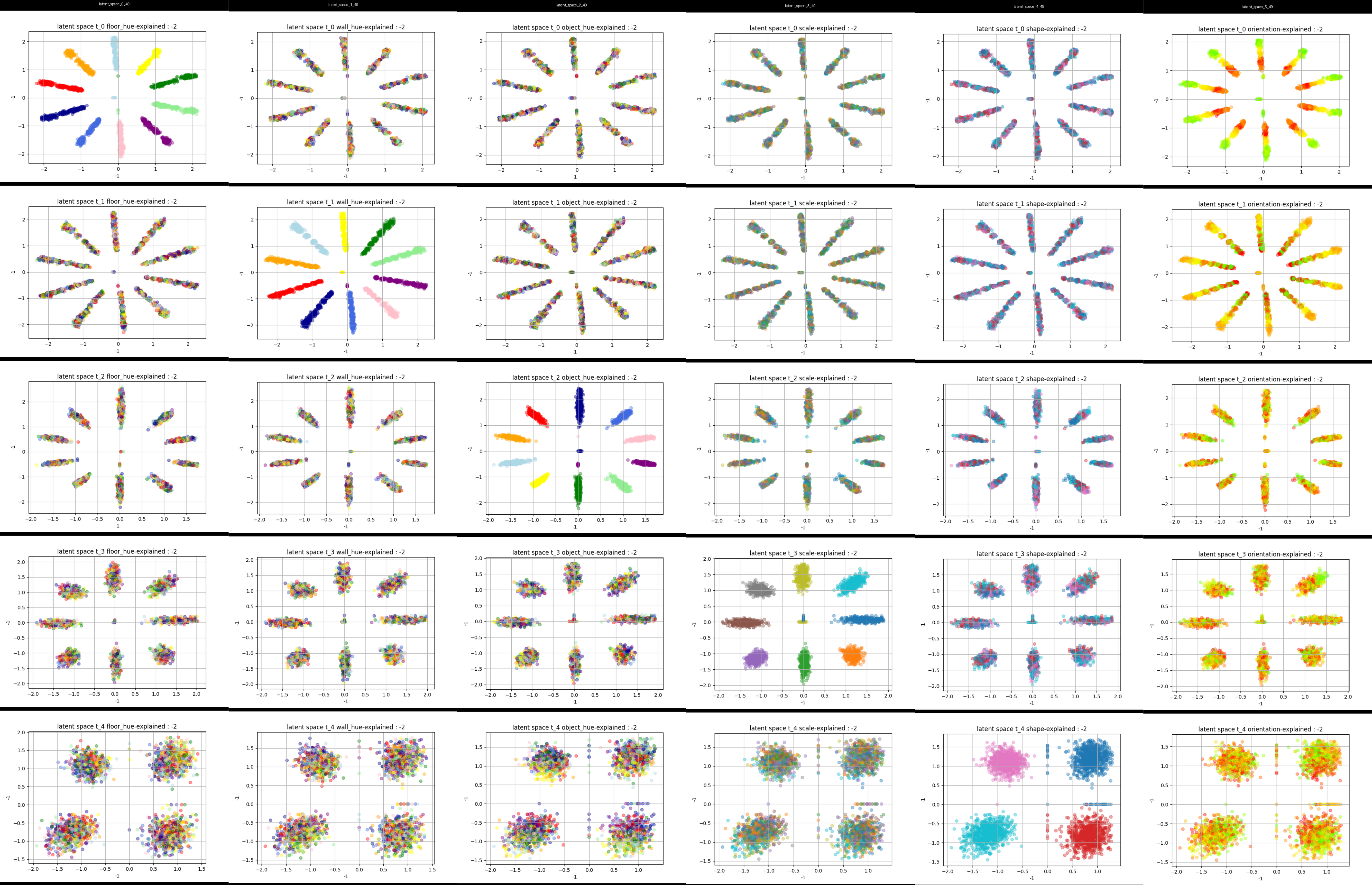}
    \caption{\footnotesize \textbf{Learned latents of Shapes3D when the $S$ space is ablated}. Rows: encoding space (top to bottom: $(\mathcal{T}_0, \mathcal{T}_1, \mathcal{T}_2, \mathcal{T}_3, \mathcal{T}_4)$). Columns: color being used, corresponding to the different possible values of a single factor (left to right: (floor hue, wall hue, object hue, scale, shape, orientation), only the coloring changes between columns).}
    \label{fig:supp_no_s}
\end{figure*}

\paragraph{No $\mathcal{S}$}
In \cref{fig:supp_no_s} we show the latents obtained by ablating the $S$ space. We can see that all $T$ encodings are well disentangled but for the first one (1st row), where the orientation factor ended up being encoded (last column). This validates that our method "learns to use" the $\mathcal{S}$ space when available.


\paragraph{Empty $\mathcal{S}$}
We also tried to encode each factor in $\mathcal{T} = \bigoplus \mathcal{T}_i$ and to leave $\mathcal{S}$ empty (but not \textit{removing} it like in the previous ablation).

We report the metrics in \cref{tab:metrics_bt1_bs100_all} for the "empty $\mathcal{S}$" configuration.
\begin{table}[h]
    \centering
    \footnotesize
    \setlength{\tabcolsep}{3pt}
    \caption{Disentanglement metrics; each labeled factor is encoded in a dedicated subspace $\mathcal{T}_i$.}
    \begin{tabular}{lcccc}
        \toprule
        \textbf{Dataset} & \textbf{D} & \textbf{C} & \textbf{I} & \textbf{FactorVAE} \\
        \midrule
        3DShapes         & $0.999994$ & $0.887436$ & $0.999995$ & $1.0000$           \\
        dSprites         & $0.907830$ & $0.740193$ & $0.933329$ & $0.9999$           \\
        MPI3D            & $0.976327$ & $0.882258$ & $0.994443$ & $0.9417$           \\
        \bottomrule
    \end{tabular}
    \label{tab:metrics_bt1_bs100_all}
\end{table}

\subsection{Additional plots}

\subsubsection{Latent spaces \hyperref[sec:latent_plots]{$\upuparrows$}}
\label{subsec:sup_latents}

\begin{figure*}
    \centering
    \includegraphics[width=\linewidth]{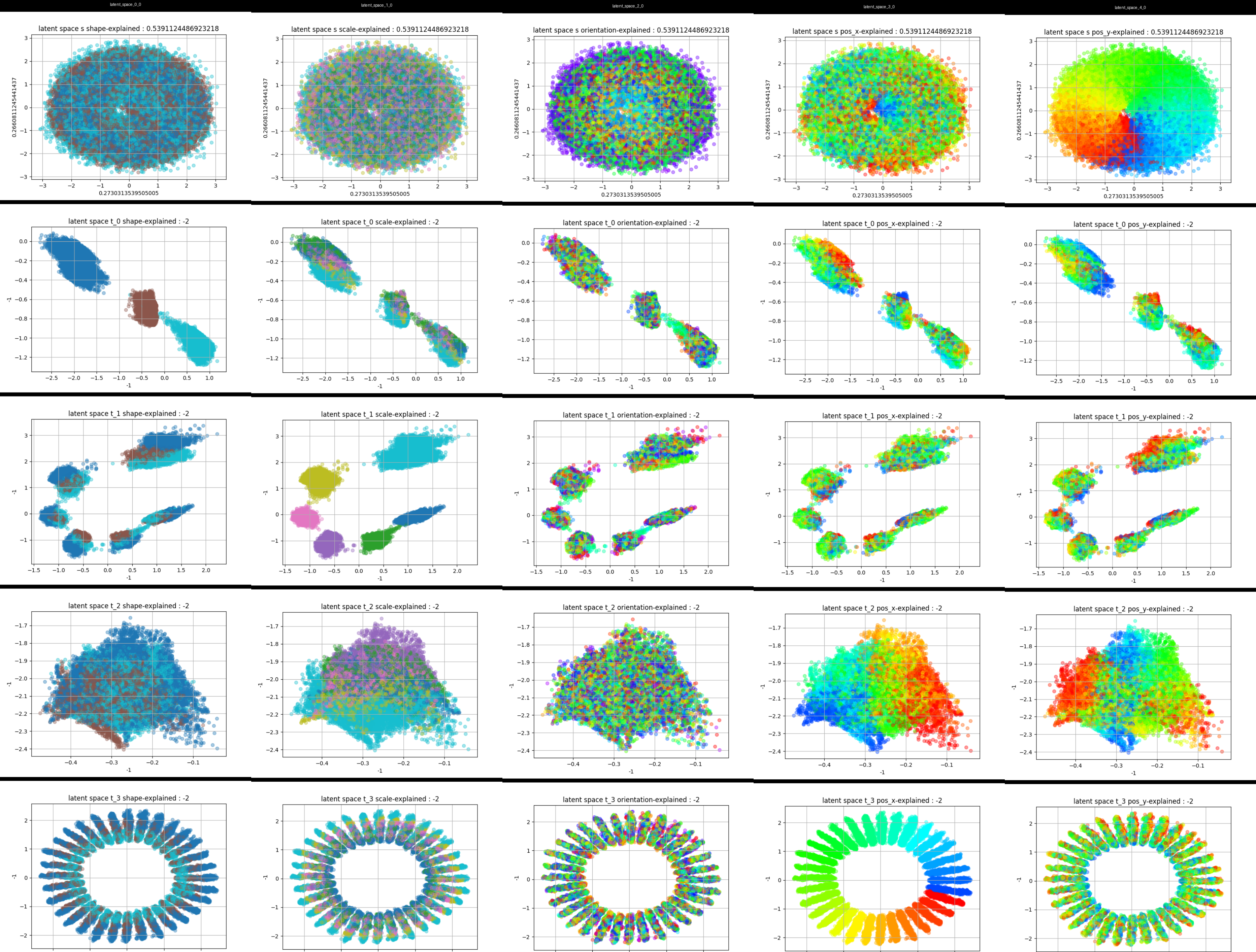}
    \caption{\footnotesize \textbf{Learned latents on dSprites}. Rows: encoding space (top to bottom: $(\mathcal{S}, \mathcal{T}_0, \mathcal{T}_1, \mathcal{T}_2, \mathcal{T}_3)$). Columns: color being used, corresponding to the different possible values of a single factor (left to right: (shape, scale, orientation, $x$ position, $y$ position), only the coloring changes between columns). For the $\mathcal{S}$ space the first 2 components of its PCA are shown.}
    \label{fig:supp_latents_dSprites}
\end{figure*}

\begin{figure*}
    \centering
    \includegraphics[width=\linewidth]{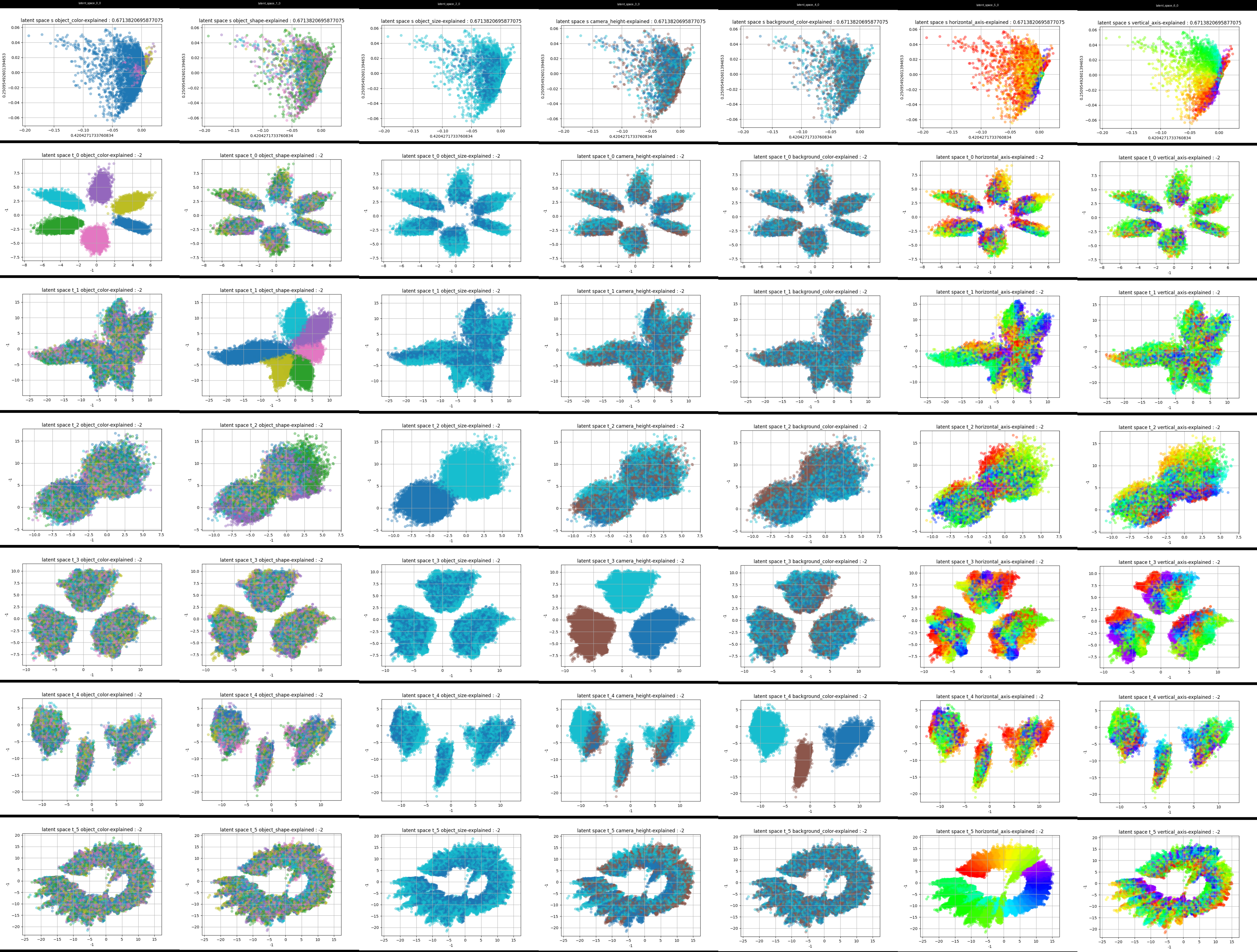}
    \caption{\footnotesize \textbf{Learned latents on MPI3D}. Rows: encoding space (top to bottom: $(\mathcal{S}, \mathcal{T}_0, \mathcal{T}_1, \mathcal{T}_2, \mathcal{T}_3, \mathcal{T}_4, \mathcal{T}_5)$). Columns: color being used, corresponding to the different possible values of a single factor (left to right: (object color, object shape, object size, camera height, background color, horizontal axis, vertical axis), only the coloring changes between columns). For the $\mathcal{S}$ space the first 2 components of its PCA are shown.}
    \label{fig:supp_latents_mpi3d}
\end{figure*}

\begin{figure*}
    \centering
    \includegraphics[width=0.8\linewidth]{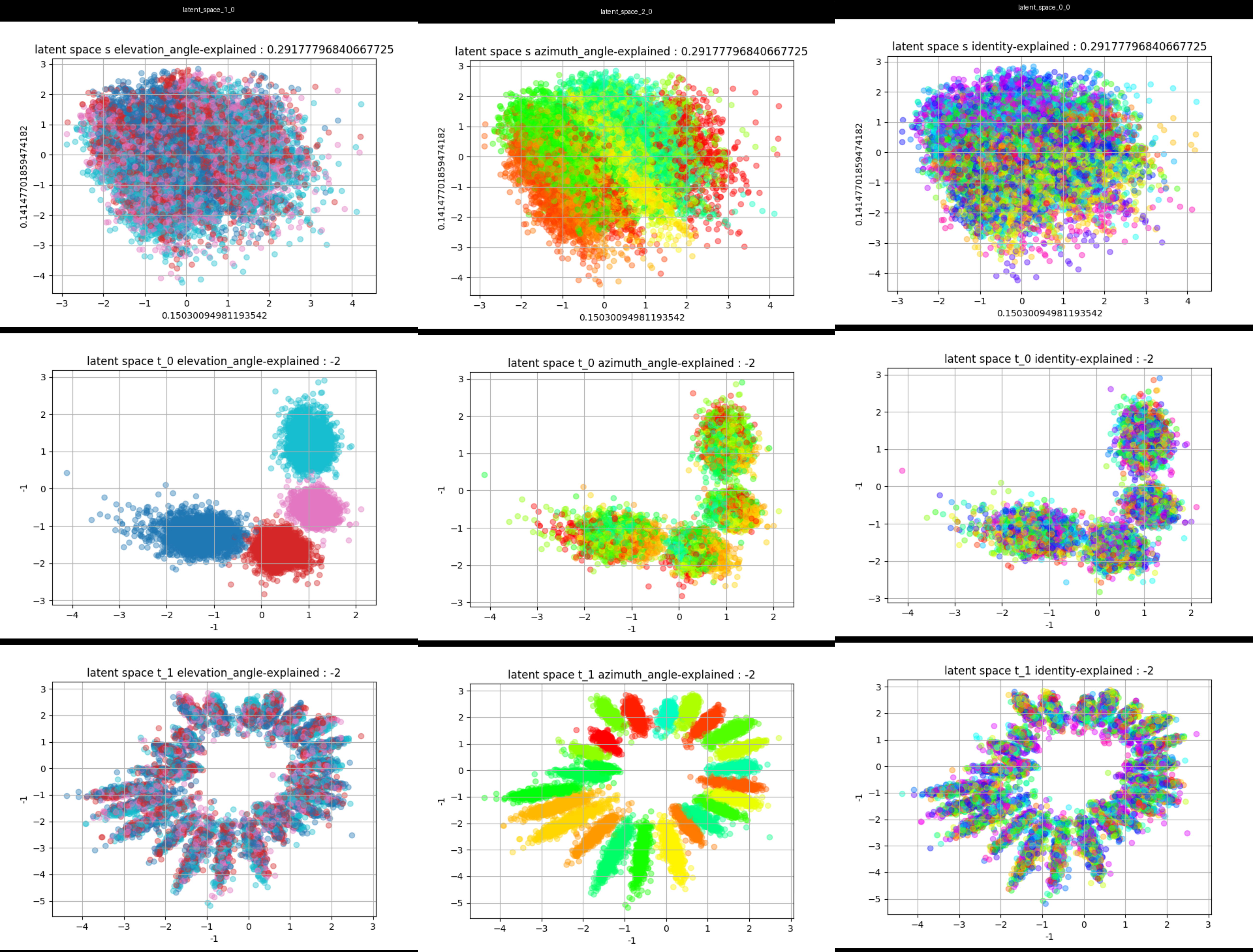}
    \caption{\footnotesize \textbf{Learned latents on cars3D}. Rows: encoding space (top to bottom: $(\mathcal{S}, \mathcal{T}_0, \mathcal{T}_1)$). Columns: color being used, corresponding to the different possible values of a single factor (left to right: (elevation angle, azimuth angle, identity), only the coloring changes between columns). For the $\mathcal{S}$ space the first 2 components of its PCA are shown.}
    \label{fig:supp_latents_cars3d}
\end{figure*}

\begin{figure*}
    \centering
    \includegraphics[width=\linewidth]{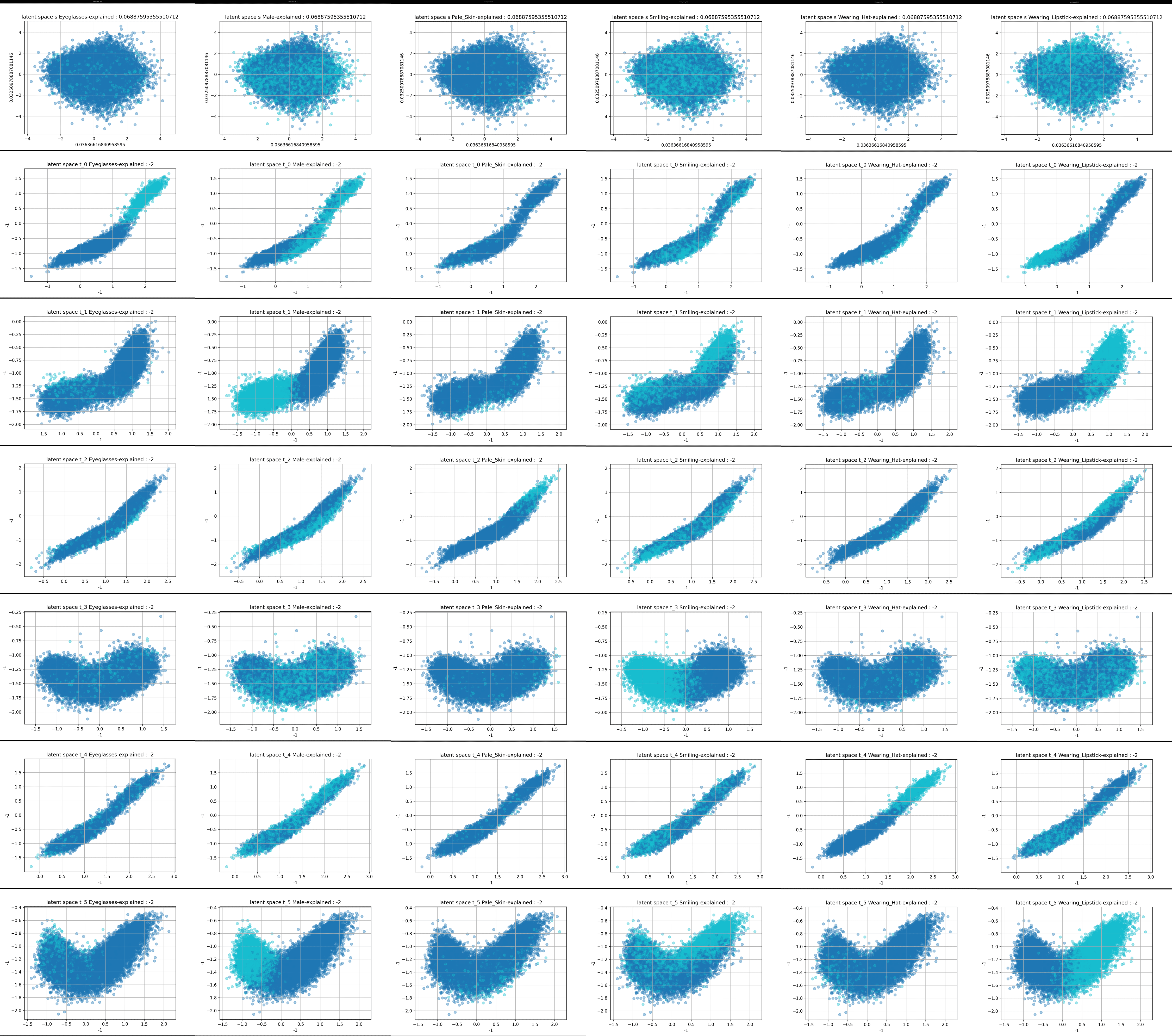}
    \caption{\footnotesize \textbf{Learned latents on CelebA}. Rows: encoding space (top to bottom: $(\mathcal{S}, \mathcal{T}_0, \mathcal{T}_1, \mathcal{T}_2, \mathcal{T}_3, \mathcal{T}_4, \mathcal{T}_5)$). Columns: color being used, corresponding to the different possible values of a single factor (left to right: (eyeglasses, male, pale skin, smiling, wearing hat, wearing lipstick), only the coloring changes between columns). For the $\mathcal{S}$ space the first 2 components of its PCA are shown. The colorings for the $\mathcal{S}$ factors are not shown as there are 34 of them.}
    \label{fig:supp_latents_celeba}
\end{figure*}

See \cref{fig:supp_no_ablation,fig:supp_latents_dSprites,fig:supp_latents_mpi3d,fig:supp_latents_cars3d,fig:supp_latents_celeba} for visualizations of the latents learned by \textsc{XFactors}. In these figures each row is a latent space, and along columns we vary the coloring according to each factor: the colors of a column match the different possible values of a single factor.\\
Strong block-specific disentanglement would mean that the subdiagonal (that contains each $\mathcal{T}_i$ spaces colored by the factor we ought to encode in it) is well structured, while all other plots appear random –to the exception of the upper right plot which might be structured somehow ($\mathcal{S}$ space colored by the factor we "left out").\footnote{Note that for CelebA we do not show the 34 colorings for the 34 left out factors...}

\subsubsection{Generation \hyperref[sec:generations]{$\upuparrows$}}
\label{sec:supp_generations}
\begin{figure*}[h]
    \centering
    \includegraphics[width=0.8\linewidth]{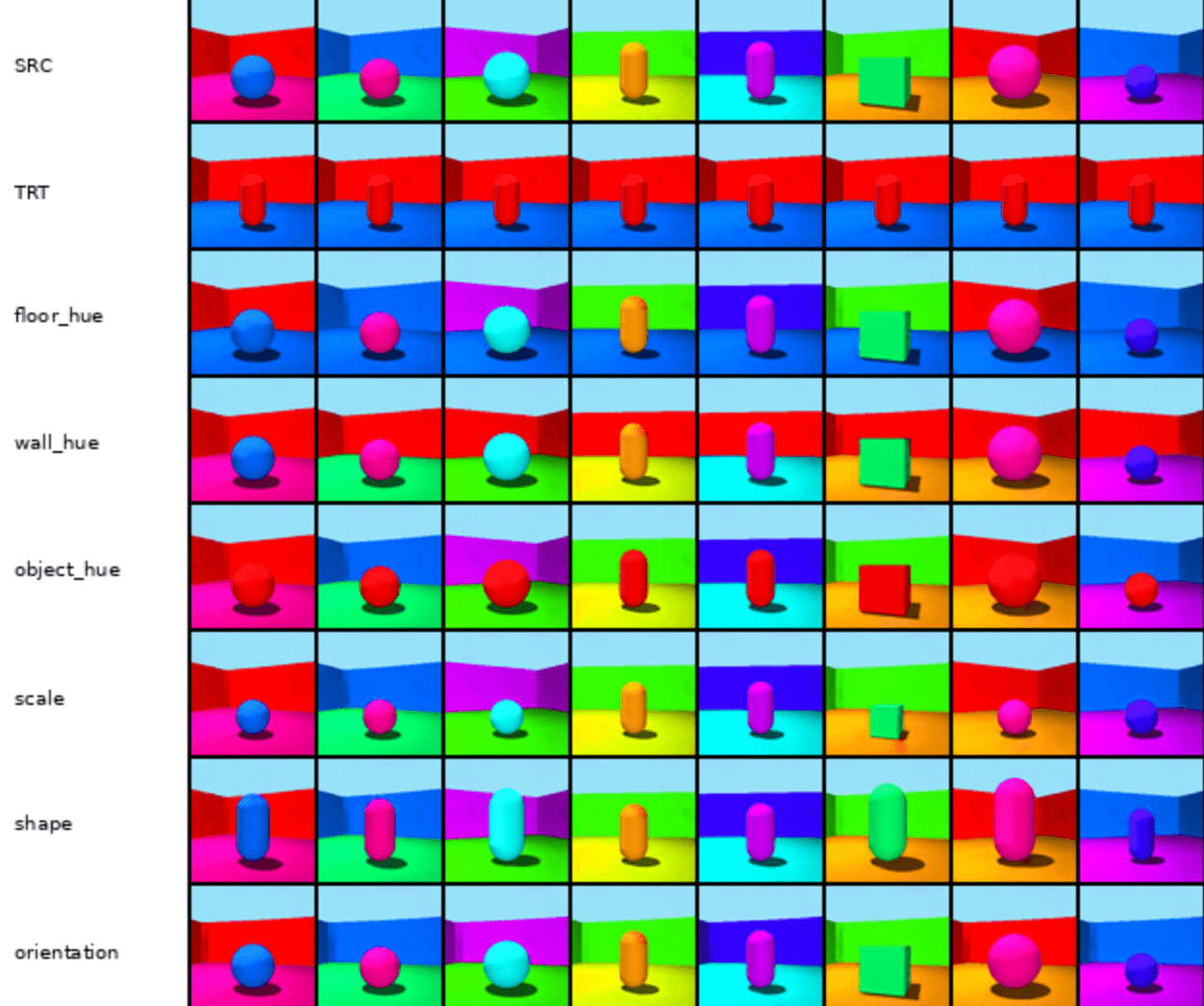}
    \caption{\textbf{Factor swapping generations on Shapes3D} For each row (source image), we replace one latent code $T_i$ with the corresponding code from the target image (shown in the header) and decode. Each column corresponds to swapping a single factor; all other latent components are kept fixed.}
    \label{fig:gen_shapes_supp}
\end{figure*}

\begin{figure*}[t]
    \centering
    \begin{subfigure}[t]{0.49\textwidth}
        \centering
        \includegraphics[width=\linewidth]{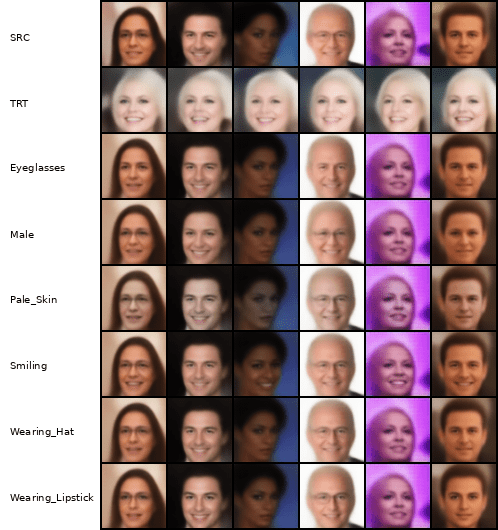}
        \caption{}
    \end{subfigure}\hfill
    \begin{subfigure}[t]{0.49\textwidth}
        \centering
        \includegraphics[width=\linewidth]{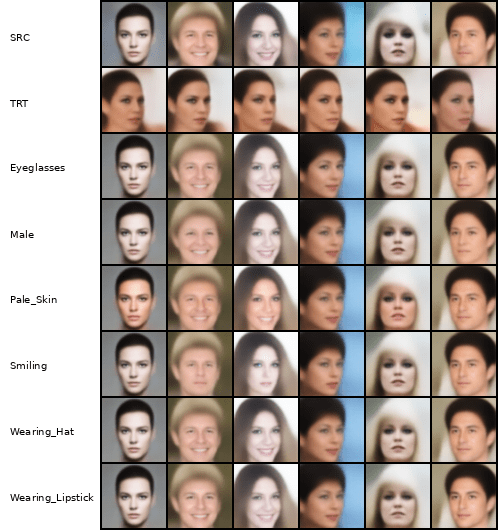}
        \caption{}
    \end{subfigure}

    \vspace{2mm}

    \begin{subfigure}[t]{0.49\textwidth}
        \centering
        \includegraphics[width=\linewidth]{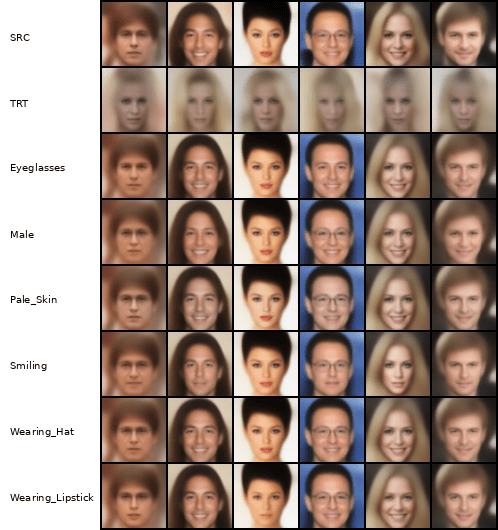}
        \caption{}
    \end{subfigure}\hfill
    \begin{subfigure}[t]{0.49\textwidth}
        \centering
        \includegraphics[width=\linewidth]{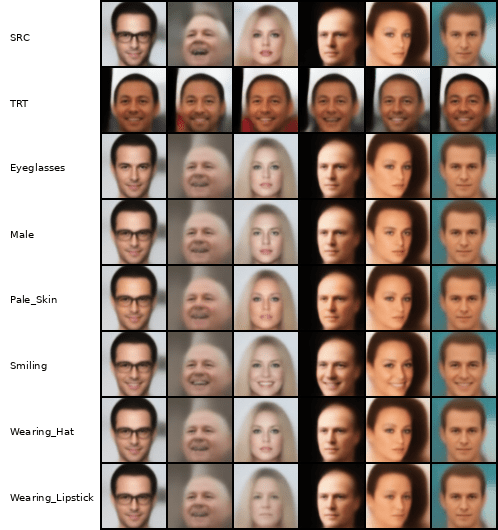}
        \caption{}
    \end{subfigure}
    \caption{\textbf{Factor swapping generations on CelebA} For each row (source image), we replace one latent code $T_i$ with the corresponding code from the target image (shown in the header) and decode. Each column corresponds to swapping a single factor; all other latent components are kept fixed.}
    \label{fig:gens_celeba_supp}
\end{figure*}

\begin{figure*}[t]
    \centering
    \begin{subfigure}[t]{0.49\textwidth}
        \centering
        \includegraphics[width=\linewidth]{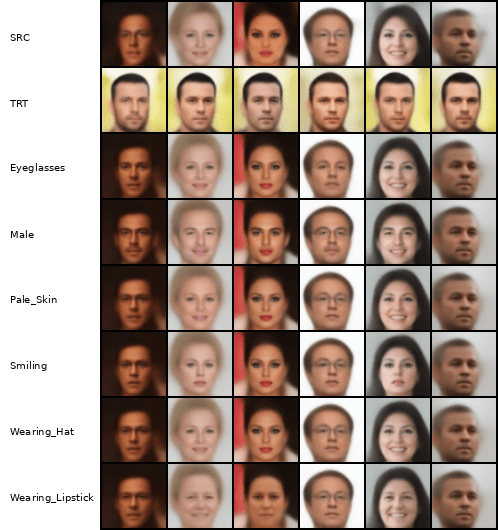}
        \caption{}
    \end{subfigure}\hfill
    \begin{subfigure}[t]{0.49\textwidth}
        \centering
        \includegraphics[width=\linewidth]{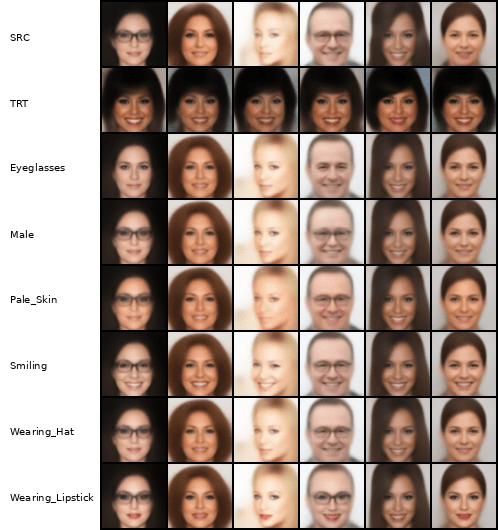}
        \caption{}
    \end{subfigure}

    \vspace{2mm}

    \begin{subfigure}[t]{0.49\textwidth}
        \centering
        \includegraphics[width=\linewidth]{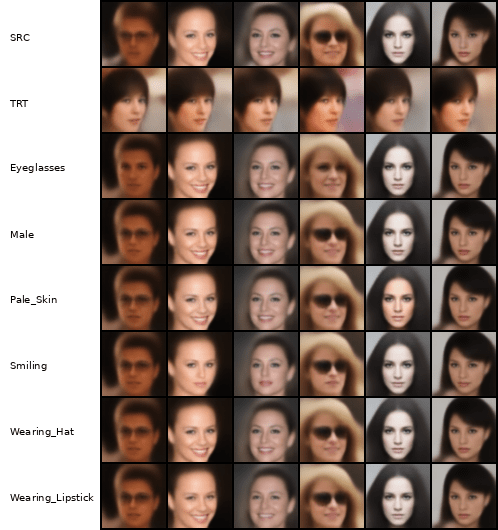}
        \caption{}
    \end{subfigure}\hfill
    \begin{subfigure}[t]{0.49\textwidth}
        \centering
        \includegraphics[width=\linewidth]{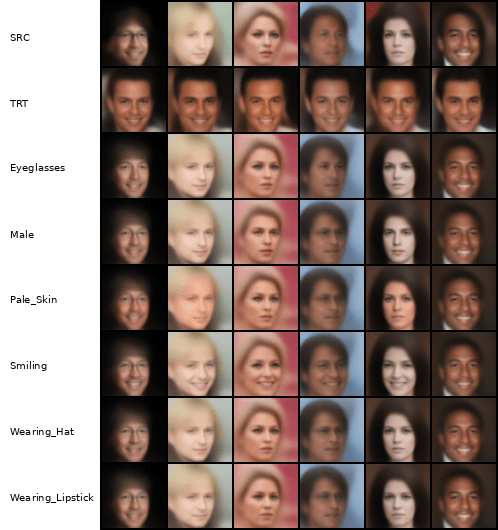}
        \caption{}
    \end{subfigure}
    \caption{\textbf{Factor swapping generations on CelebA} For each row (source image), we replace one latent code $T_i$ with the corresponding code from the target image (shown in the header) and decode. Each column corresponds to swapping a single factor; all other latent components are kept fixed.}
    \label{fig:gens_celeba_supp_1}
\end{figure*}

\begin{figure*}[t]
    \centering
    \begin{subfigure}[t]{0.49\textwidth}
        \centering
        \includegraphics[width=\linewidth]{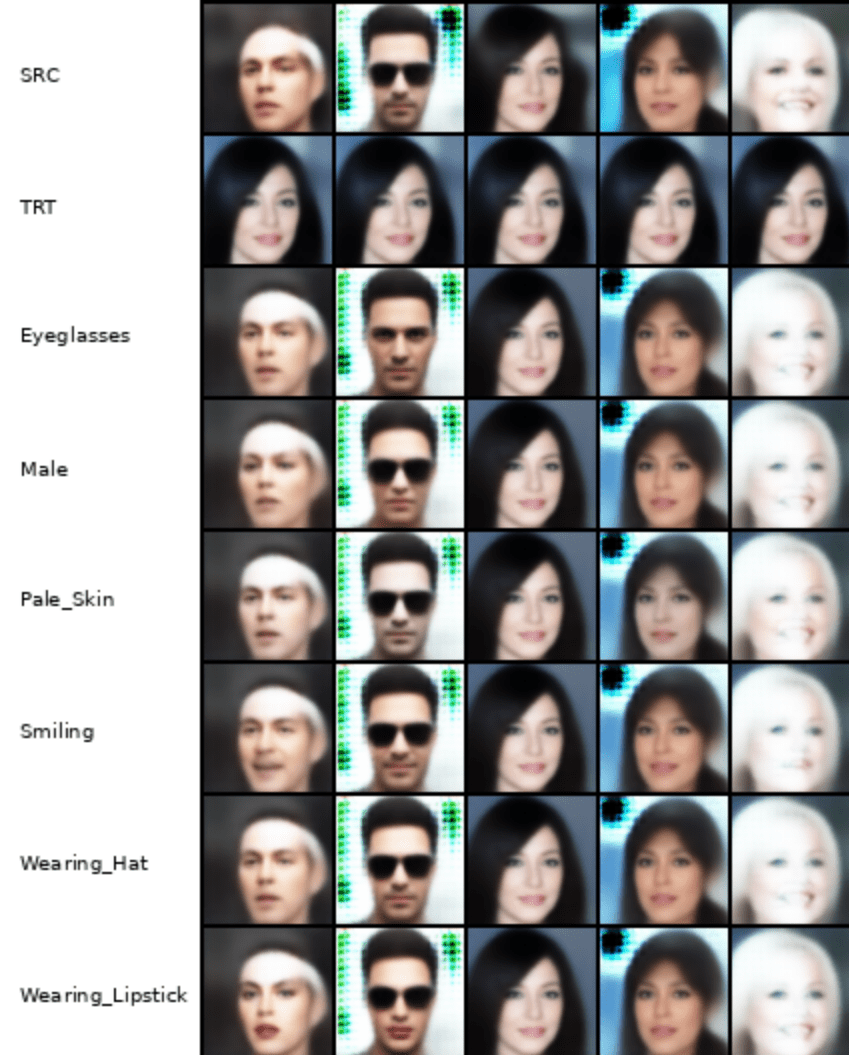}
        \caption{}
    \end{subfigure}\hfill
    \begin{subfigure}[t]{0.49\textwidth}
        \centering
        \includegraphics[width=\linewidth]{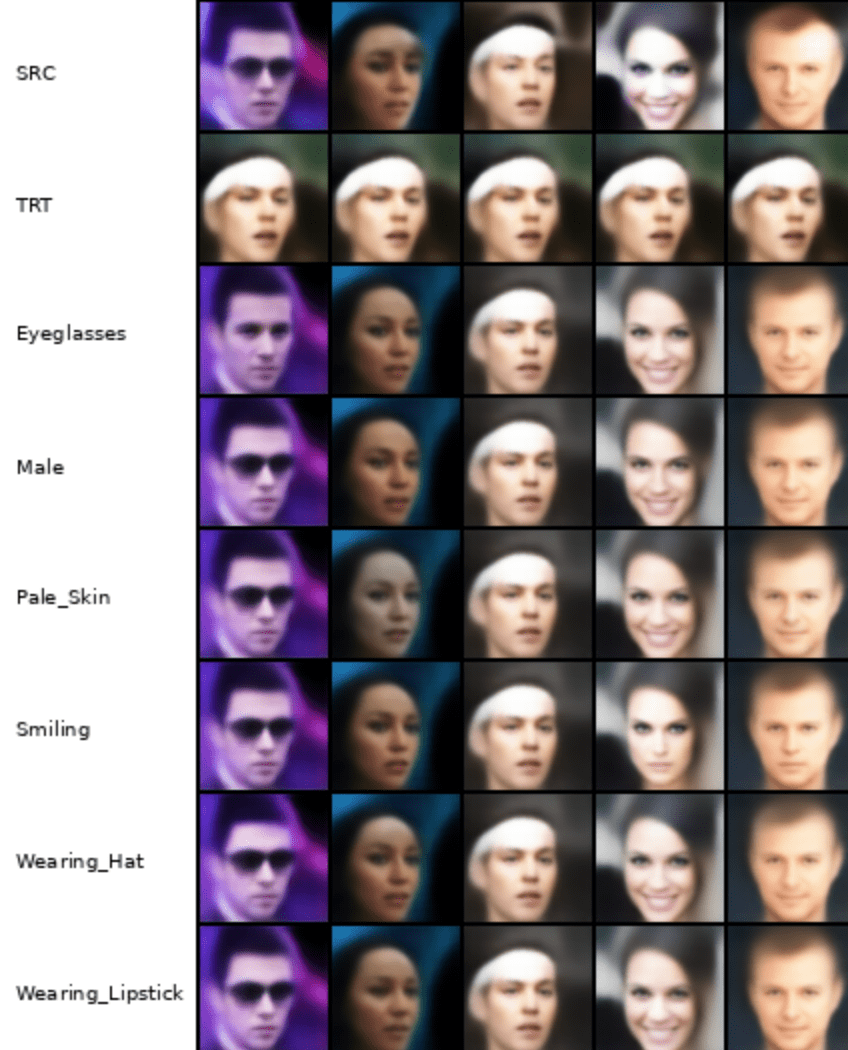}
        \caption{}
    \end{subfigure}

    \vspace{2mm}

    \begin{subfigure}[t]{0.49\textwidth}
        \centering
        \includegraphics[width=\linewidth]{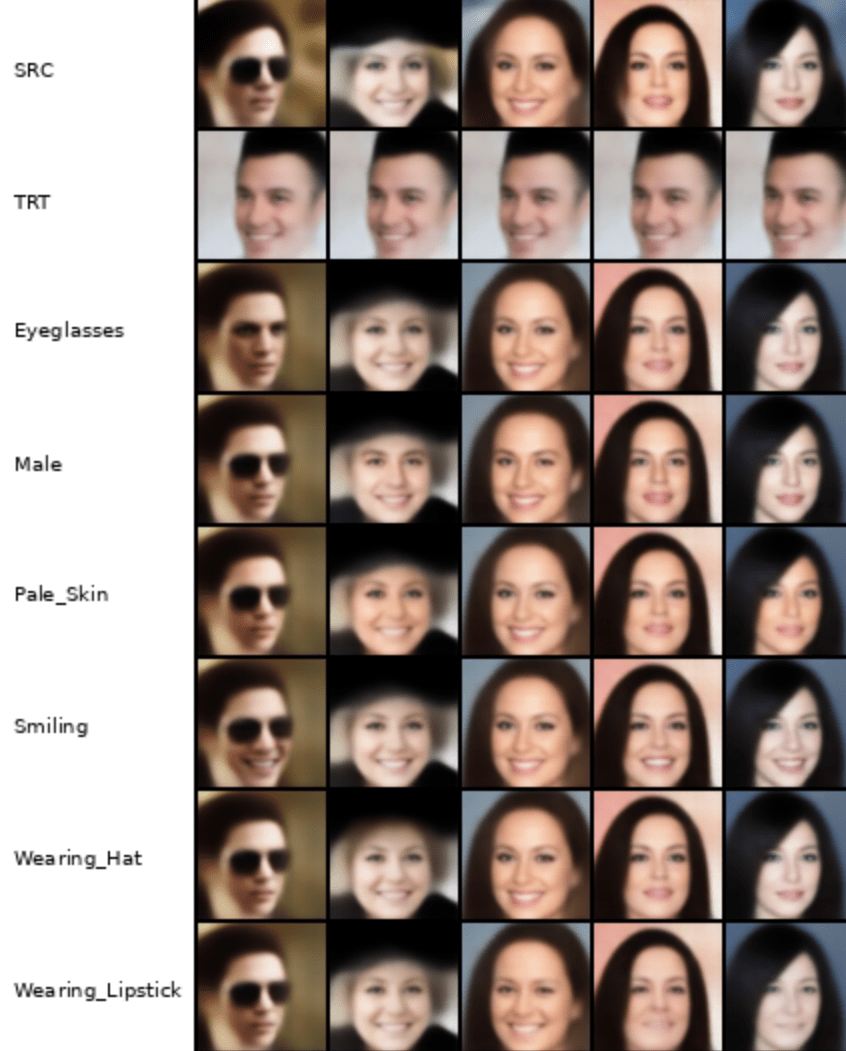}
        \caption{}
    \end{subfigure}\hfill
    \begin{subfigure}[t]{0.49\textwidth}
        \centering
        \includegraphics[width=\linewidth]{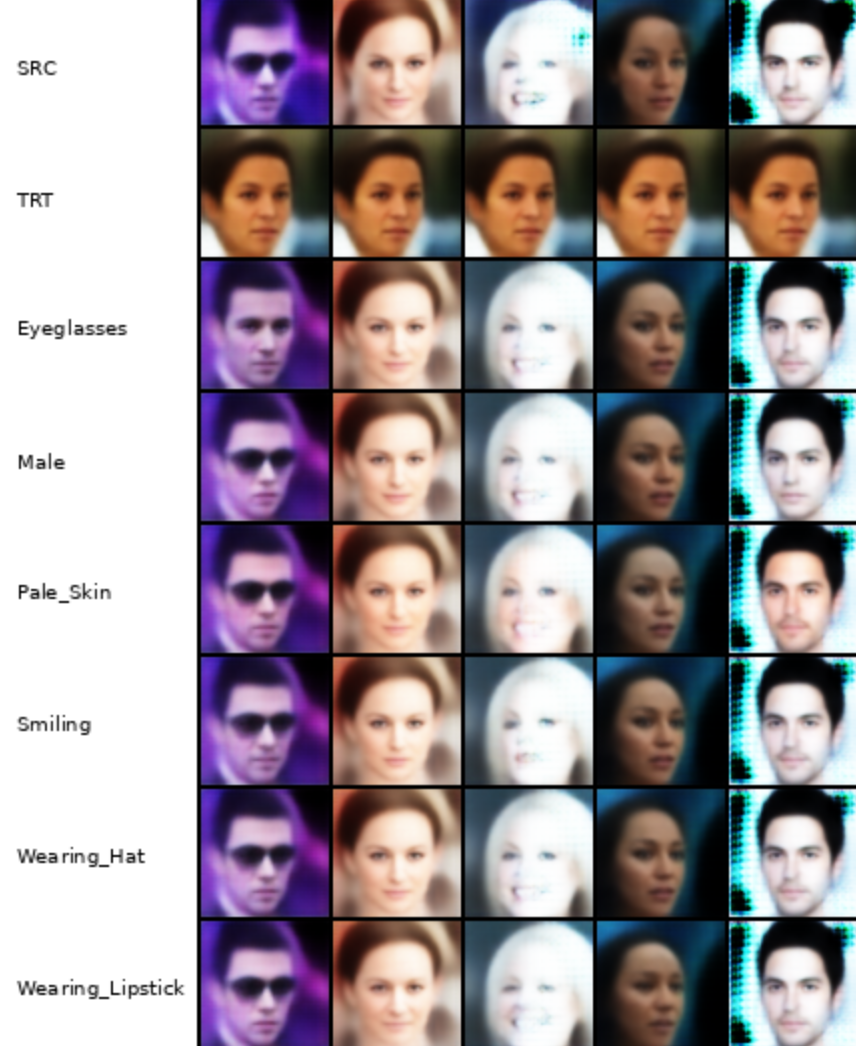}
        \caption{}
    \end{subfigure}
    \caption{\textbf{Factor swapping generations on CelebA} For each row (source image), we replace one latent code $T_i$ with the corresponding code from the target image (shown in the header) and decode. Each column corresponds to swapping a single factor; all other latent components are kept fixed.}
    \label{fig:gens_celeba_supp_2}
\end{figure*}

\begin{figure*}[h]
    \centering
    \includegraphics[width=0.6\linewidth]{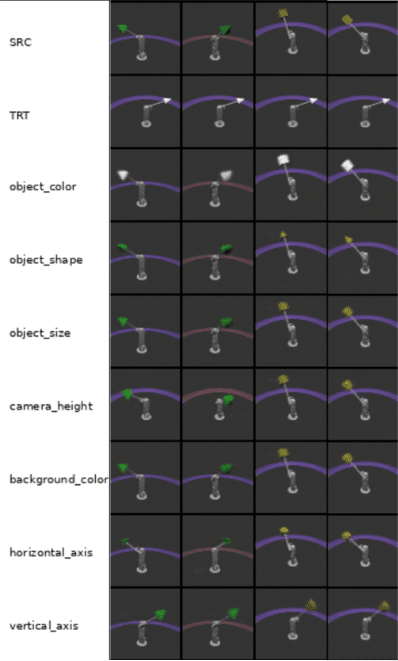}
    \caption{\textbf{Factor swapping generations on MPI3D} For each row (source image), we replace one latent code $T_i$ with the corresponding code from the target image (shown in the header) and decode. Each column corresponds to swapping a single factor; all other latent components are kept fixed.}
    \label{fig:gen_mpi_supp}
\end{figure*}

\begin{figure*}[h]
    \centering
    \includegraphics[width=\linewidth]{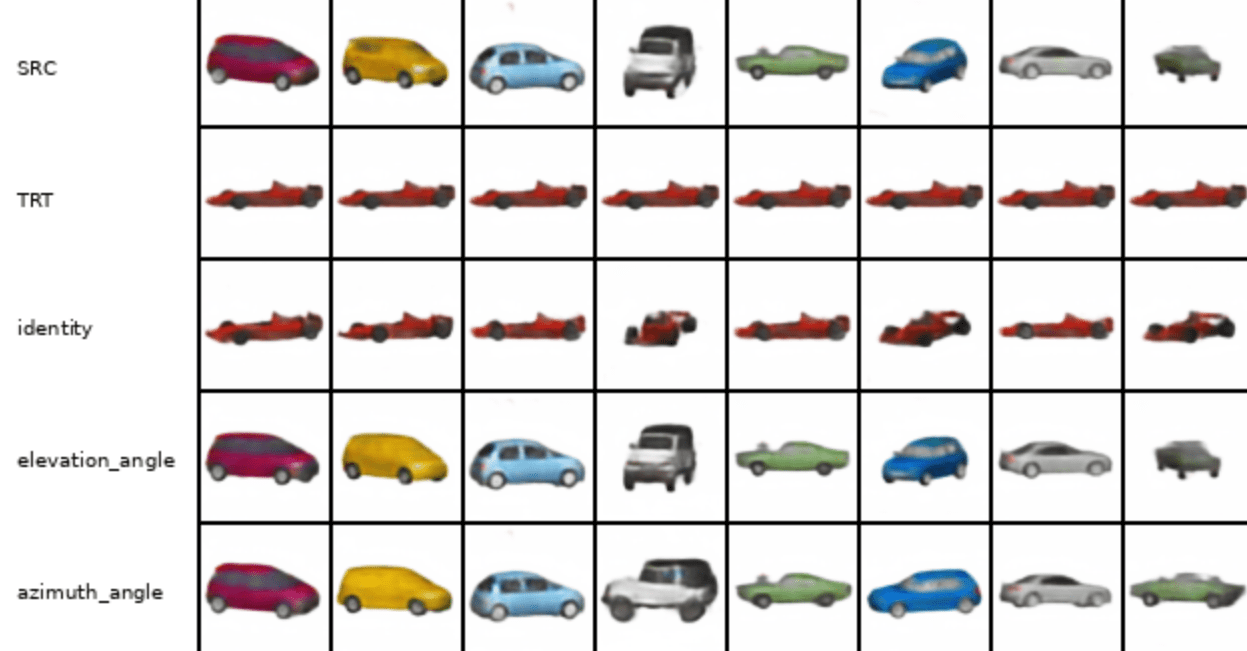}
    \caption{\textbf{Factor swapping generations on Cars3D} For each row (source image), we replace one latent code $T_i$ with the corresponding code from the target image (shown in the header) and decode. Each column corresponds to swapping a single factor; all other latent components are kept fixed.}
    \label{fig:gen_cars3d_supp}
\end{figure*}

\begin{figure*}[h]
    \centering
    \includegraphics[width=0.7\linewidth]{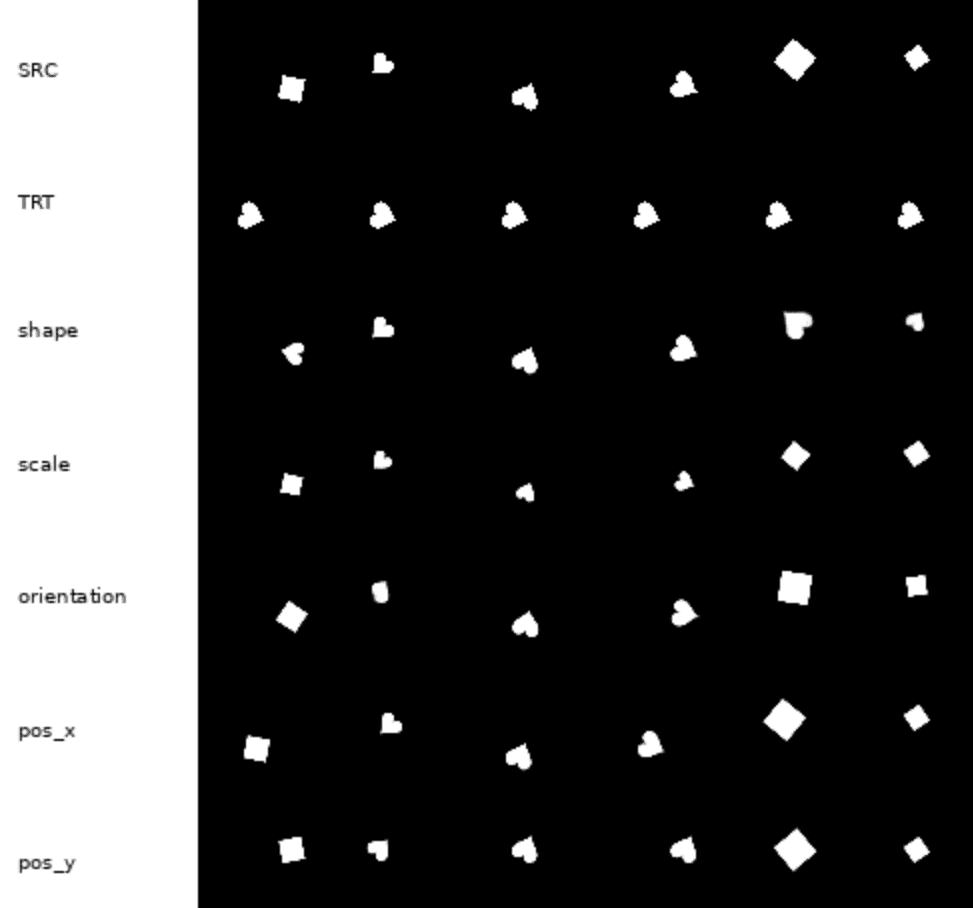}
    \caption{\textbf{Factor swapping generations on dSprites} For each row (source image), we replace one latent code $T_i$ with the corresponding code from the target image (shown in the header) and decode. Each column corresponds to swapping a single factor; all other latent components are kept fixed.}
    \label{fig:gen_dsprites_supp}
\end{figure*}

We present in \cref{fig:gen_shapes_supp,fig:gens_celeba_supp,fig:gens_celeba_supp_1,fig:gens_celeba_supp_2,fig:gen_mpi_supp,fig:gen_cars3d_supp,fig:gen_dsprites_supp} additional factor swapping generations.

\clearpage

\end{document}